\documentclass[journal,twoside]{IEEEtran}

\usepackage{enumerate}
\usepackage{mathtools}
\usepackage{microtype}

\usepackage{pdflscape}

\usepackage{tabularx}
\usepackage{multirow}
\usepackage{textcomp}
\usepackage{csquotes}
\usepackage{booktabs}

\usepackage{array}

\usepackage{xr-hyper}
\usepackage{hyperref}
\usepackage{amsmath}
\usepackage{amssymb}
\usepackage{lscape}
\usepackage{lineno}
\usepackage{array}
\usepackage{float}
\usepackage{todonotes}
\usepackage{tabu}
\usepackage{siunitx}
\usepackage{longtable}
\usepackage{pdflscape}
\usepackage{setspace}
\usepackage{textcomp}
\usepackage{rotating}
\usepackage{xcolor}
\usepackage{acronym}
\usepackage{multicol}
\usepackage{xtab}
\usepackage{multirow}
\usepackage{lipsum} % For generating dummy text

\usepackage{xspace}
\usepackage{acronym} 
\usepackage{tablefootnote}
\usepackage{multicol}
\usepackage{float}

\usepackage{amssymb}% http://ctan.org/pkg/amssymb
\usepackage{pifont}% http://ctan.org/pkg/pifont

\DeclareUnicodeCharacter{2212}{-}

\usepackage{comment}
\def\BibTeX{{\rm B\kern-.05em{\sc i\kern-.025em b}\kern-.08em
    T\kern-.1667em\lower.7ex\hbox{E}\kern-.125emX}}

\ifCLASSINFOpdf
\else
 
\fi

\usepackage[a4paper, total={180mm,257mm}]{geometry}

\begin{document}

%
% paper title
% can use linebreaks \\ within to get better formatting as desired
%\title{Perspectives in hybrid edge-cloud computing: a case study on energy efficiency in buildings}
%\title{A hybrid tow-stage edge-based anomaly detection of building energy consumption}

%\title{Anomaly-Net: Deep Anomaly Detection of Building Energy Consumption Using Time-Series Imaging}

\title{Federated Learning for Computer Vision}

% author names and affiliations
% use a multiple column layout for up to three different
% affiliations

\author{\IEEEauthorblockN{Yassine Himeur\IEEEauthorrefmark{1},
Iraklis Varlamis\IEEEauthorrefmark{2}, 
Hamza Kheddar\IEEEauthorrefmark{3}, 
Abbes Amira\IEEEauthorrefmark{4,5},
Shadi Atalla\IEEEauthorrefmark{1},
Yashbir Singh\IEEEauthorrefmark{6},
Faycal Bensaali\IEEEauthorrefmark{3} and
Wathiq Mansoor\IEEEauthorrefmark{1}
}\\
\IEEEauthorblockA{\IEEEauthorrefmark{1}
College of Engineering and Information Technology, University of Dubai, Dubai, UAE}\\
\IEEEauthorblockA{\IEEEauthorrefmark{2}Department of Informatics and Telematics, Harokopio University of Athens, Greece}\\
\IEEEauthorblockA{\IEEEauthorrefmark{3}LSEA Laboratory, Electrical Engineering Department, University of Medea, Algeria}\\
\IEEEauthorblockA{\IEEEauthorrefmark{4}Department of Computer Science, University of Sharjah, UAE}\\
\IEEEauthorblockA{\IEEEauthorrefmark{5}Institute of Artificial Intelligence, De Montfort University, Leicester, United Kingdom}\\
\IEEEauthorblockA{\IEEEauthorrefmark{6}Department of Radiology, Mayo Clinic, Rochester, MN, USA}\\
\IEEEauthorblockA{\IEEEauthorrefmark{6}Electrical Engineering Department, Qatar University, Qatar}\\
}

% use for special paper notices
%\IEEEspecialpapernotice{(Invited Paper)}

% make the title area
\maketitle

\begin{abstract}
Computer Vision (CV) is playing a significant role in transforming society by utilizing machine learning (ML) tools for a wide range of tasks. However, the need for large-scale datasets to train ML models creates challenges for centralized ML algorithms. The massive computation loads required for processing and the potential privacy risks associated with storing and processing data on central cloud servers put these algorithms under severe strain. To address these issues, federated learning (FL) has emerged as a promising solution, allowing privacy preservation by training models locally and exchanging them to improve overall performance. Additionally, the computational load is distributed across multiple clients, reducing the burden on central servers. 
This paper presents, to the best of the authors' knowledge, the first review discussing recent advancements of FL in CV applications, comparing them to conventional centralized training paradigms. It provides an overview of current FL applications in various CV tasks, emphasizing the advantages of FL and the challenges of implementing it in CV. To facilitate this, the paper proposes a taxonomy of FL techniques in CV, outlining their applications and security threats. It also discusses privacy concerns related to implementing blockchain in FL schemes for CV tasks and summarizes existing privacy preservation methods. Moving on, the paper identifies open research challenges and potential future research directions to further exploit the potential of FL and blockchain in CV applications.
\end{abstract}

\begin{IEEEkeywords}
Computer vision (CV), Federated learning (FL), Blockchain, Horizontal federated learning (HFL), Vertical federated learning (VFL).
\end{IEEEkeywords}

\IEEEpeerreviewmaketitle

\section{Introduction}
In recent years, the emergence and evolution of machine learning (ML) has carved out a distinct niche for itself within the broader realm of artificial intelligence (AI). Specifically, ML emphasizes the design, development, and training of algorithms and software agents to sift through, interpret, and make data-driven decisions based on patterns hidden within vast datasets \cite{himeur2021artificial}. These artificial agents are increasingly being equipped with multifaceted sensory capabilities that allow them to interact more intuitively with their environment. Among these, vision stands out as the most profound, granting machines the ability to 'see' and 'understand' the world around them in ways comparable to human cognition \cite{sayed2022artificial}.
Given the intrinsic importance of visual data, computer vision (CV) has rapidly risen to prominence within the ML community. CV seeks to enable machines to interpret and derive meaningful conclusions from visual data, mimicking the human ability to recognize, process, and respond to visual stimuli \cite{ng2021federated}. This capability not only empowers artificial agents to better grasp their surroundings but also paves the way for more informed decision-making processes \cite{al2023automated}. By incorporating CV techniques, artificial agents can move beyond mere data analysis, delving into a more profound comprehension of the visual realm \cite{sayed2023time}. The integration of such insights holds the potential to revolutionize various domains by enhancing the depth of understanding and the efficacy of decision-making processes in AI systems \cite{teng2023data,himeur2022deep}.

%Recently, machine learning (ML) has been viewed as a sub-field of artificial intelligence (AI) focusing on the construction and training of software agents that analyse data and take rational decisions in specific tasks \cite{himeur2021artificial}. Artificial agents have access to a wide range of sensory modalities, with vision being the most useful of all \cite{sayed2022artificial}. As a result, computer vision (CV) has become a crucial ML task that may assist artificial agents in making wise judgments, by providing them with a thorough perception of what they see and or their surroundings \cite{ng2021federated,himeur2022deep}.

Understanding the contents and concepts of an image involves a significant amount of information which is connected with image segmentation, extraction of features and objects, and synthesis of the scene as a whole \cite{elmir2023ecg}. Although the CV and perception task is reflexively performed by humans, grace to their ability for abstraction, it is still quite complex for artificial agents \cite{chouchane2023improving}. It is also hard for humans to explain how they perceive the world and develop ML modules that behave in a similar manner \cite{kheddar2023deep}.  
The four main techniques that are employed for training machines to perform CV tasks are either based on statistics (i.e. on patterns learned from large training datasets), on the logic expressed in the form of rules, on deep neural networks (DNNs) that capture the non-linear relations between image features and the final decision or on genetic and evolutionary algorithms that combine multiple decisions in order to find the one that maximizes the overall performance \cite{himeur2023face,esposito2001machine}.

% \textcolor{black}{b.	Computation and privacy challenges of CV }

The huge advancements and impressive results in the performance of CV algorithms in tasks such as image classification, image segmentation, object detection, and scene perception, came from a shift from signal processing methods to solutions that rely on deep learning (DL) methods \cite{copiaco2023innovative}. The biggest challenge for DL methods is that they require huge computational resources and energy for training, which in turn makes them hardly applicable in many application setups, where decisions have to be taken on the edge, using low resources and limited power (e.g. in drones, mobile phones, etc.) \cite{alyamkin2019low}. 
For this purpose low-power CV challenges have been established and new energy-efficient algorithms and light-weight DL architectures have been proposed, such as reservoir computing \cite{tong2018reservoir}, or random neural networks \cite{mohamed2002study}.

Another challenge for CV systems is the protection of user privacy.
Collecting millions of people's images and videos poses serious privacy risks, which must be seriously considered \cite{himeur2022latest,quach2022digital}. First of all, the right to be forgotten is constantly violated by companies that collect visual data and retain it permanently or use it for multiple purposes. Also, through numerous flaws, such delicate visual data could be exploited or exposed \cite{himeur2018robust}. Secondly, the privacy-preserving analysis of images and videos from video surveillance applications, where CV methods are used to detect violations (e.g. mask-wearing, distance keeping, etc.) in public spaces can be quite challenging too \cite{patrikar2022anomaly,himeur2018robust}. Such videos frequently include inadvertently caught private objects including faces, car plates, computer screens, and more. The ability of methods to detect and even identify humans and other objects in unconstrained environments can put at risk the anonymity of people in monitored places and, if not used properly, can become a threat to citizen privacy \cite{xiang2022being}.

% \textcolor{black}{c. Edge computing, FL, split learning and beyond for CV applications}

The rise of edge computing architectures has introduced a new potential to privacy-preserving CV, since with the proper use of the limited edge resources, it has become possible to perform basic CV tasks without transferring sensitive data to the cloud \cite{alsalemi2022innovative}.
Sharing the trained model rather than releasing the actual data also helps to retain the privacy of the data, without losing on prediction performance \cite{sayed2021intelligent}.
Due to their built-in privacy-preserving features, federated learning (FL) \cite{yang2019federated} and split learning (SL) \cite{gupta2018distributed} are the two ML techniques, that perform on visual data on a distributed manner and have attracted the interest of researchers in the CV field. Such approaches assume a client-server (edge-cloud) architecture, where the clients usually have fewer resources than the server. The clients train their models individually, using their own data, and then exchange their models either with the server or among them in order to synchronize what they learned \cite{khan2021federated}. The ML models are trained locally on the client (edge) devices and this happens in parallel, as long as the clients receive and process training data. Periodically, they exchange models and aggregate them following simple or more sophisticated strategies, which may involve additional training on the cloud for avoiding the bias introduced in the clients \cite{giorgas2020online}. In the case of SL the training of the model layers is also split between the edge and the cloud. Only a few layers are trained on the edge, in order to facilitate training with little resources, and the remaining layers are trained on the cloud \cite{abuadbba2020can}.

A risk that emerges from this distributed training process for CV models is the exposure to malicious users that intentionally introduce noise or falsified models in order to bias the FL model for their benefit \cite{jere2020taxonomy}. Adversarial training techniques can be employed to strengthen the FL models, but even adversarial training has leakages \cite{zhang2022privacy}. The use of blockchain technologies can help mitigate many of the threats of data or model-sharing methods. All the model-sharing activities of a node that is allowed to share are stored as transactions in the blockchain and all information about the providers' profiles is also stored in the blockchain \cite{himeur2022blockchain}. In this multi-party setup, when a node needs to update its model by using the models of other nodes, first performs a request for models to its neighbouring nodes, then validates the nodes by checking the blockchain and retrieves their models from the blockchain. The resulting FL model is consequently stored on the blockchain in a new transaction \cite{lu2019blockchain}.

\subsection{Survey Methodology}

%\subsection{Research Questions}
FL is emerging as a powerful machine learning paradigm that allows for the training of AI models across numerous decentralized devices while maintaining data privacy. In essence, FL enables devices to collaboratively learn a shared model without having to share raw data, a feature that holds immense potential for applications in diverse domains, particularly in CV. In CV, the applications of FL range from object detection and image classification to semantic segmentation, among others, with an array of potential real-world use cases in sectors such as healthcare, autonomous driving, and surveillance systems. However, the adoption of FL in CV is not without its challenges. Issues related to model performance, communication efficiency, data heterogeneity, and privacy preservation need to be addressed and carefully managed. This systematic literature review aims to delve deep into these challenges, exploring the progress, limitations, and future directions of applying FL in the realm of CV. The review will address several research questions that encapsulate the core facets of this intriguing intersection of FL and CV, which can summarized in Table \ref{RQs}.

\begin{table*}[t!]
\caption{Research questions covered in this review.}
\label{RQs}
\small
\begin{tabular}{
m{8mm}
m{60mm}
m{100mm}
}
\hline

RQ\# & Question	& Objective   \\

\hline

RQ1 & What is the primary motivation behind conducting a review of FL for CV? 	& 
Present the underlying reasons or driving factors for undertaking a comprehensive review specifically on the application of FL in the domain of CV.    \\ \hline

RQ2 & How is the problem of data privacy handled in FL for CV?	&   Investigate the mechanisms, strategies, or solutions implemented within the FL framework, specifically for CV tasks, to address and ensure data privacy.    \\ \hline

RQ3 & How is model performance measured and optimized in FL scenarios? 	& Delve into the methods, metrics, and techniques used to evaluate the efficacy of models trained using FL.    \\ \hline

RQ4 & How are issues related to data distribution across nodes (non-IID data) managed?	& Understand the strategies, techniques, and solutions employed to handle the challenges posed by non-Independent and Identically Distributed (non-IID) data in decentralized or FL environments.    \\ \hline

RQ5 & What are the strategies to handle communication efficiency in FL?	& Discuss the various techniques and methods implemented to optimize or enhance the communication process between nodes or devices in a FL system.   \\ \hline

RQ6 & What are the approaches to ensure robustness and security in FL? 	& Investigate the various methodologies, techniques, or safeguards that are employed to protect FL systems against potential vulnerabilities, attacks, or failures.  \\ \hline

RQ7 & How are FL systems for CV deployed in the real world?	& Explore and understand the practical applications, implementation challenges, infrastructure requirements, and real-world scenarios where FL techniques are used specifically in the domain of CV.    \\ \hline

RQ8 & What are the challenges and limitations in applying FL to CV tasks?	& Identify the potential difficulties, drawbacks, and constraints encountered when using FL in the context of CV.    \\ \hline

RQ9 & What are the trends and future directions in FL for CV?	& Explore the current trajectories, innovations, and potential advancements in the application of FL to CV.    \\ \hline

\hline

\end{tabular}
\end{table*}

%\subsection{Inclusion/exclusion Strategy}
During our systematic review, we initially collected \textcolor{black}{912} papers. To ensure the relevance of the literature, we applied basic criteria such as title, abstract, and topic alignment with our research question. We then established detailed inclusion and exclusion criteria to streamline the selection process. The inclusion criteria encompassed papers proposing FL solutions, discussing the applications of FL in CV, implementing techniques, or proposing enhanced versions of FL. Conversely, the exclusion criteria were applied to exclude publications that did not specifically use FL, it was only mentioned in the literature review , used for comparison purposes or used other research sectors, etc.
One author took the lead in the selection strategy and conducted the initial screening, ensuring consistency with our research theme. 
% Any disagreements regarding the suitability of specific works were resolved through discussions with other authors. 
After removing duplicates, we identified \textcolor{black}{385} unique articles. We then conducted a thorough assessment of the remaining articles by carefully reviewing titles, abstracts, and conclusions. Based on this assessment, we narrowed down the selection to \textcolor{black}{255} articles that exhibited relevance based on title and abstract. In the subsequent stage, we applied the specified inclusion and exclusion criteria to the remaining articles, leading to the exclusion of certain studies that did not meet our criteria. 

During the selection process, the following exclusion criteria were adopted: (i) duplicate records, (ii) 
papers that did not comment on the performance of FL in CV, (iii) papers related to the implementation of similar but not FL techniques, (iv) papers related to research sectors other than CV, and (v) papers written in languages other than English.
Furthermore, the following inclusion criteria were used to select relevant literature: (i) proposes an improvement an FL-based solution in CV, (ii) addresses the privacy and security issues in CV using FL techniques, (iii) measured and optimized model performance in FL, (iv) discusses the deployment of FL systems in real-world CV applications

By applying these exclusion and inclusion criteria, we ensured that the selected articles provided insights, solutions, or advancements specifically related to the filter bubble phenomenon in RSss. This process resulted in a final selection of 28 articles that met our inclusion criteria. In order to ensure a comprehensive review, we conducted a reference scan of the selected articles, which led us to identify an additional 6 relevant papers. Consequently, a total of 34 articles were included in our systematic review on the existence of the filter bubble. %Figure \ref{fig:selection} provides an overview of our research selection criteria and the distribution of publications obtained from each database.

\subsection{Related reviews}

%Only a few recent survey articles have addressed the topic of FL. In the following discussion, the outline of the primary resemblances and distinctions between our review and the existing surveys on FL are summarized in Table \ref{tab:RW}.

While several studies have recently addressed the topic of FL and its various applications, no work has specifically focused on reviewing the applications of FL in CV. For instance, \cite{wahab2021federated} discusses various applications of FL in communication and networking systems, including wireless sensor networks, IoT, and vehicular networks, and highlight some challenges that need to be addressed to fully realize the potential of FL, such as security concerns, communication overheads, and model heterogeneity. Moving on, \cite{li2021survey} provides an extensive examination of FL while reviewing the growing concerns about data privacy and security, (ii) offers design considerations and solutions; (iii) Highlights different privacy-preserving approaches utilized in FL; and (iv) it suggests potential future paths. Similarly, \cite{aledhari2020federated} discusses (i) the key enabling technologies and protocols used in FL such as secure aggregation, differential privacy, homomorphic encryption, etc.; and (ii) challenges faced by FL such as communication overheads, heterogeneity of devices and data, privacy concerns, and security issues.

Besides, \cite{nguyen2021federated} provides a thorough exploration of the increasing applications of FL in IoT networks. It begins by acknowledging the limitations of centralized AI methods reliant on data collection and privacy concerns. FL, as a collaborative and distributed AI approach, offers a solution for intelligent IoT applications without data-sharing needs. The authors showcase various use cases in healthcare, transportation, etc., where FL can be utilized in IoT networks. Additionally, they address the challenges and potential opportunities linked with FL.
Nguyen et al. \cite{nguyen2021federated} discuss IoT-based FL and its applications and summarizes the technical aspects, key contributions, and limitations of each reference work. Moreover, \cite{abdulrahman2020survey} introduces a novel classification of FL subjects and research domains and encompasses comprehensive taxonomies that address various challenging aspects, contributions, and trends in the literature. Additionally, the survey delves into significant challenges and potential research paths for developing more robust FL systems.

%\textcolor{black}{Contribution of the review}
\subsection{Contribution }
By contrast to the above-mentioned reviews, this study presents to the best of the authors' knowledge the first review of FL-based contributions in CV. Typically, it (i) offers a comparative perspective by presenting related reviews, thereby situating its contributions in the broader scholarly conversation; (ii) provides foundational knowledge about FL, including its definition and problem formulation; (iii) discusses different aggregation methods, covering averaging aggregation, progressive Fourier aggregation, and FedGKT aggregation, (iv) presents a detailed overview of privacy technologies associated with FL, including the Secure MPC model, differential privacy, and homomorphic encryption, (v) distinguishes and elaborates on supervised, unsupervised, and semi-supervised FL and evaluates CV schemes based on FL.

Moreover, it offers an in-depth look at how FL is employed in various CV tasks such as object detection, video surveillance, healthcare, autonomous driving, and more.
Moving on, it addresses significant challenges and concerns in the FL domain, such as communication overhead, device compatibility, and human bias. Additionally, by pinpointing the specific obstacles that researchers and practitioners might face, the review contributes to the growing body of knowledge about the intersection of FL and CV.
Lastly, beyond its retrospective lens, the review also peers into the future, suggesting potential avenues of research and development in the field.

%The subsequent sections of this review are structured as follows: Section \ref{background} provides basic background information on FL models and their use in CV tasks. Then, section \ref{applications} performs an overview of the main applications of FL in CV, which reveals the limits and risks of  FL schemes. Section \ref{issues} discusses the main open issues and illustrates the current mitigation strategies. Under the aforementioned prism, this section emphasizes the challenges that arise from the use of FL in CV. Section \ref{future} focuses on the future research directions relevant to the role of FL and Blockchain in CV, and finally Section \ref{conclusion} concludes the paper with the main findings of this work.

\section{Background of FL}
\label{background}
FL enables the training of AI models without the sharing of training data.
% , which offers an approach to unlock data to feed new AI applications. 
While most AI has been trained on data gathered and crunched in a unique repository, today's ML algorithms are shifting towards a decentralized scheme.  ML models can collaboratively be trained on edge devices, e.g.,  mobile phones, laptops, or private servers.

\subsection{Definition}

%, as portrayed in Fig. \ref{fig:HFL}. 

By overviewing recent studies of FL frameworks proposed in CV applications, FL models can be categorized into
three types: \\

\begin{figure*}[ht!]
\begin{center}
\includegraphics[width=0.9\linewidth]{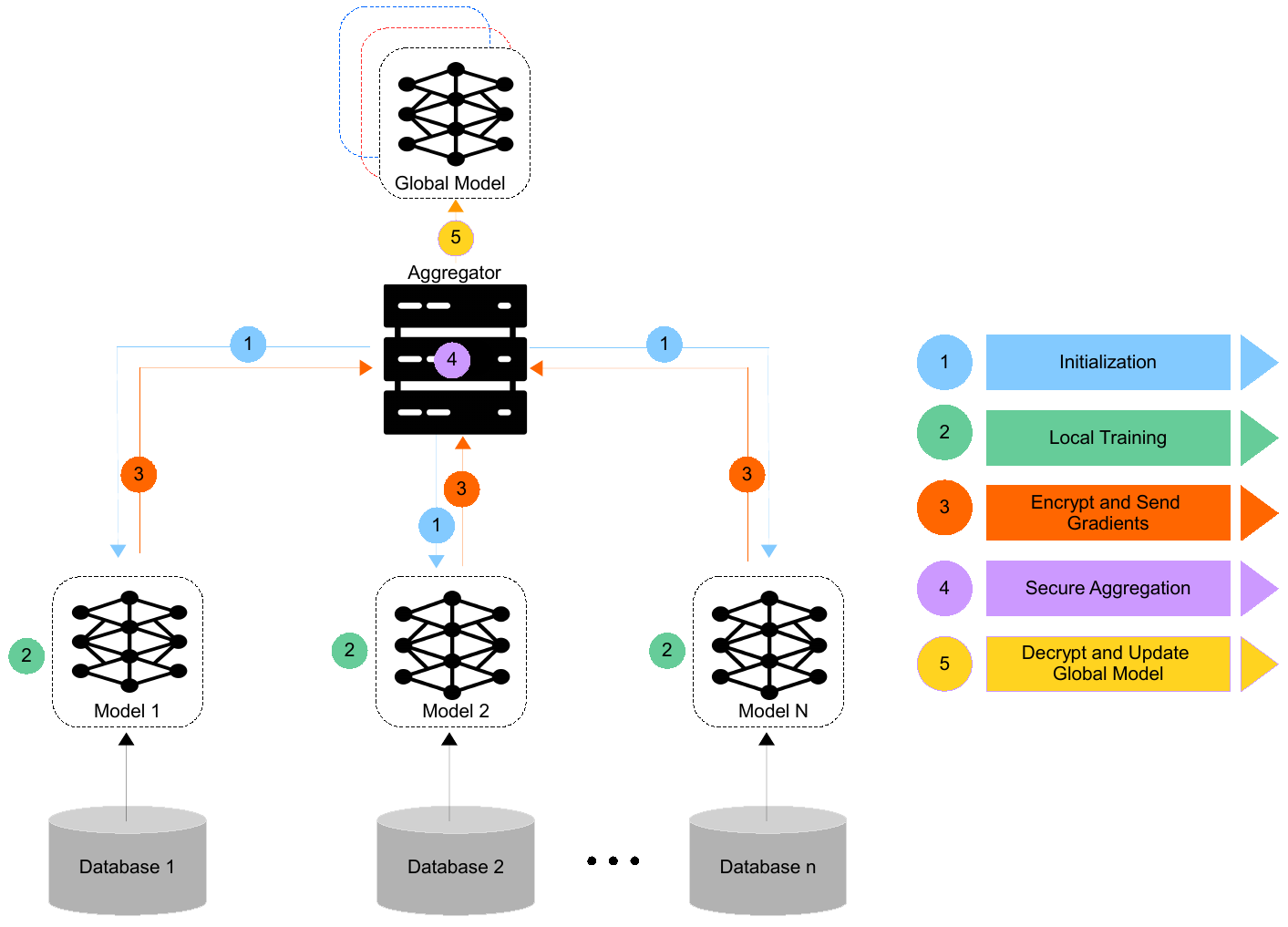}\\
\end{center}
\caption{Architecture of HFL used in CV.}
\label{fig:HFL}
\end{figure*}

\noindent\textbf{Horizontal FL (HFL):} refers to training a shared overall model by the clients using their datasets, which are characterized by the same feature space but have different sample spaces, as portrayed in Fig. \ref{fig:HFL}. 
In this regard, local FL participants can adopt
the same AI model (e.g., neural network-NN) for training their datasets. Afterwards, the
server combines the local updates communicated by the local clients to develop the overall update without accessing local data \cite{cheng2020federated}. 

Let us consider a HFL scenario with $N$ parties. Each party $i$, where $i \in \{1, 2, \dots, N\}$, has a local dataset denoted by $D_i$. Each $D_i$ consists of a set of input data points $X_i$ and corresponding labels $Y_i$. The goal is to train a global model denoted by $f$, which can generalize well on all parties' datasets. The model $f$ is typically represented as a parameterized function, such as a neural network, with trainable parameters denoted by $\theta$. The training process in HFL typically involves the following steps: (1) the process starts with the initialization of local model parameters $\theta_i$ for each participating party $i$. (2) Then, each party performs local training on its own dataset $D_i$ using its local model parameters $\theta_i$. During local training, a local loss function $L_i(\theta_i)$ is optimized to minimize the discrepancy between the local model's predictions and the corresponding labels in $D_i$. The aim is to update the local model parameters $\theta_i$ and reduce the local loss $L_i$. (3) After local training, the parties encrypt their gradients calculated with respect to their local loss functions. The encrypted gradients ensure that the parties' local updates remain private. The encrypted gradients are then sent to a trusted aggregator or a central server for further processing. (4) The trusted aggregator or central server performs secure aggregation on the encrypted gradients received from the parties. Secure aggregation techniques, such as secure multiparty computation (MPC) or homomorphic encryption, are employed to aggregate the gradients while maintaining privacy. The aggregator can compute the average of the encrypted gradients or perform more sophisticated aggregation schemes, such as averaging the parameters, of $\theta_i$ to obtain the global model parameters $\theta$. (5) Once the secure aggregation is completed, the aggregated gradients are decrypted by the trusted aggregator or central server. The global model parameters $\theta$ are then updated using the decrypted aggregated gradients. This update ensures that the global model benefits from the collective knowledge of all parties while preserving the privacy of individual data.
The mathematical formulation of HFL objective typically involves minimizing a global loss function $L(\theta)$ over all parties' datasets. This can be expressed as:

\begin{equation}
\label{theta}
 L(\theta) = \sum_{i=1}^{N} L_i(\theta_i) + \lambda.R(\theta),   
\end{equation}

\noindent where $\sum_{i=1}^{N} L_i(\theta_i)$ represents the sum of local loss functions for all parties, $\lambda$ is a regularization parameter, and $R(\theta)$ represents a regularization term that helps prevent overfitting and encourages model simplicity. \\

\begin{figure*}[ht!]
\begin{center}
\includegraphics[scale=0.65]{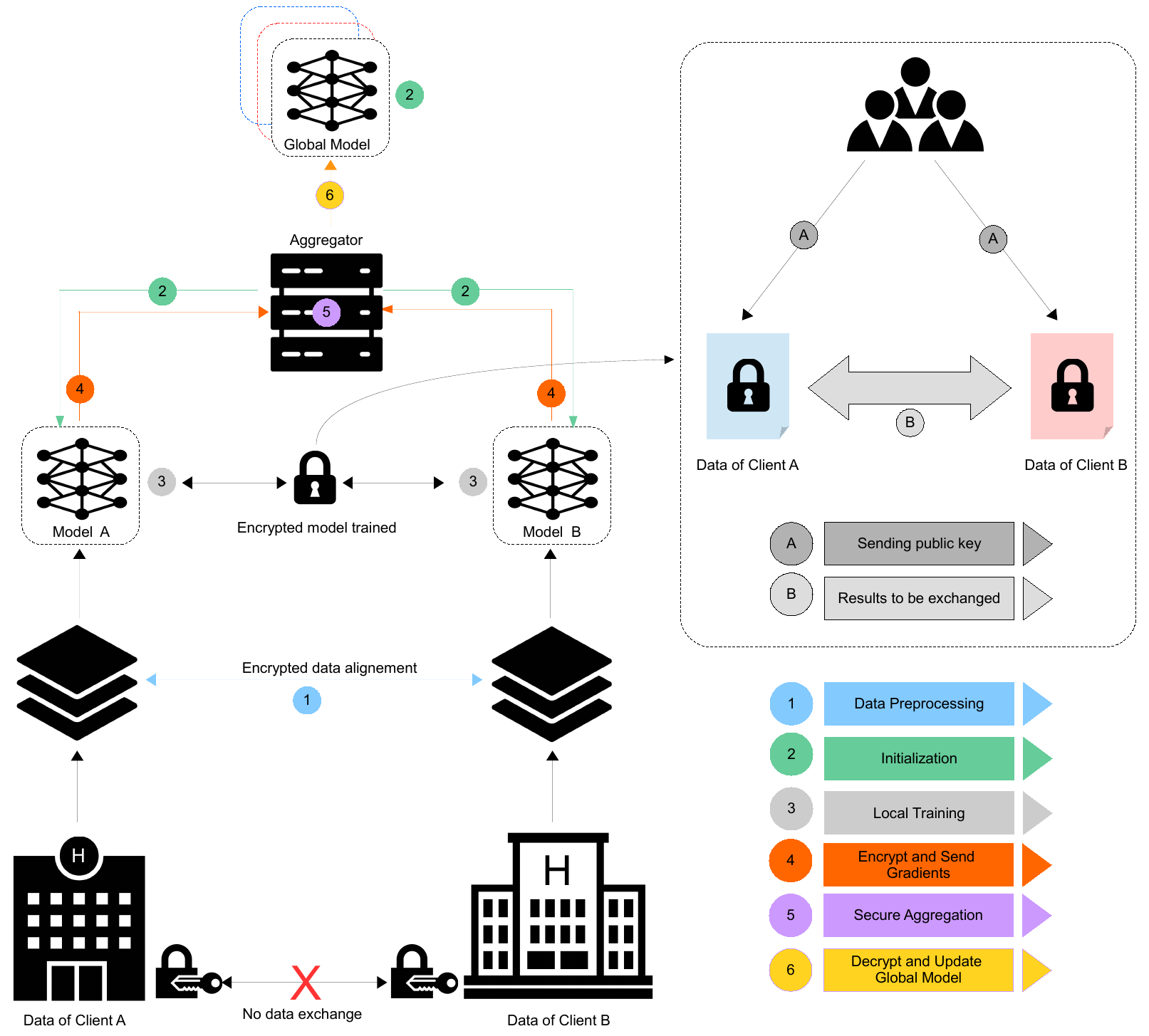}\\
\end{center}
\caption{Architecture of VFL used in CV.}
\label{fig:VHL}
\end{figure*}

\noindent\textbf{Vertical FL (VFL):} as illustrated in Fig. \ref{fig:VHL}, VFL trains ML models on datasets having the same sample space, but different feature spaces. In this context, techniques that rely on entity alignment can be used in combination with encryption to overcome the problem of data sample overlapping at distributed clients during the local training \cite{yang2019parallel}. With VFL different clients can cooperatively train their models without sharing their data (that differ in features), by only sharing their predictions for the samples they have in common. Let us consider a scenario with two parties, party $A$ and party $B$. Party $A$ holds a dataset with features $X_A$ and labels $Y_A$, while party $B$ holds a dataset with features $X_B$ and labels $Y_B$. The goal is to jointly train a model that leverages the complementary information from both parties' datasets without sharing their raw data. The training process in VFL typically involves the following steps: (1) The datasets at party $A$ and party $B$ need to be pre-processed to ensure compatibilities and feature alignment, such as data transformation, feature mapping, etc. (2) Initialize the model parameters $\theta_A$, $\theta_B$, and $\theta$ of the local and global models. (3) Each party independently performs local training on its own dataset using its local model parameters $\theta_A$, and $\theta_B$. Party $A$ uses $X_A$ and $Y_A$, while party $B$ uses $X_B$ and $Y_B$. Local loss functions $L_A(\theta)$ and $L_B(\theta)$ are optimized to update the local model parameters and reduce the discrepancies between the local model's predictions and the corresponding labels in each party's dataset. The steps (4), (5), and (6) are similar to steps (3), (4), and (5) of HFL. The mathematical formulation of $ L(\theta) $ in VFL, for $N$ parties, remains the same as the HFL case (Equation \ref{theta}). \\

%
%An example of VFL in IMoT applications can be the shared learning model among entities in a smart healthcare environment, e.g., hospitals and an insurance company. In this context, a hospital and an insurance company (different data feature), which serve patients (same sample space) can join a VFL process to cooperatively train an AI model using their datasets, e.g., historical medical records at hospitals and healthcare costs at the insurance company for intelligent healthcare decision making.

\begin{figure*}[ht!]
\begin{center}
\includegraphics[scale=0.7]{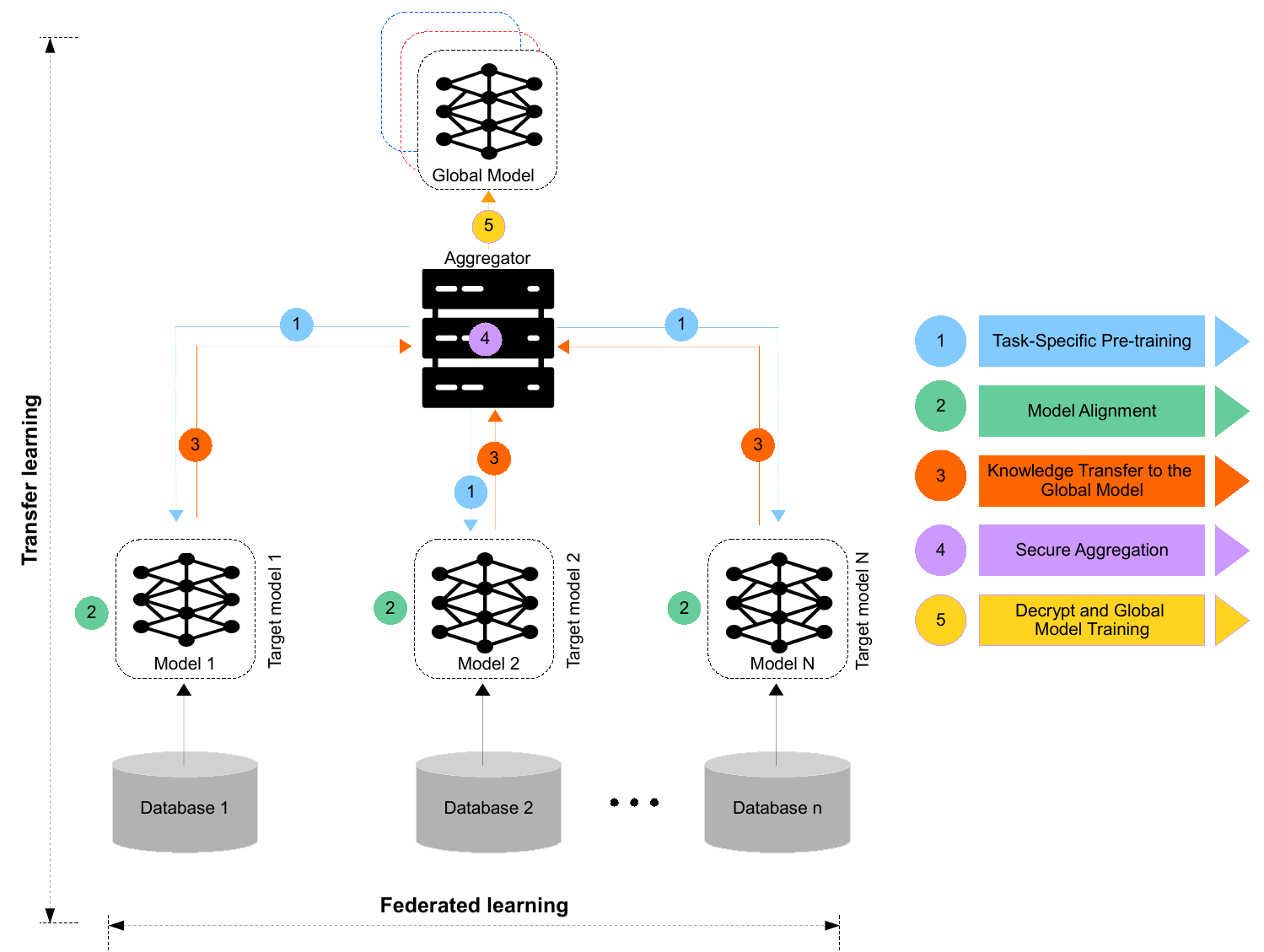}\\
\end{center}
\caption{Architecture of FTL used in CV.}
\label{fig:FTL}
\end{figure*}

\noindent\textbf{Federated transfer learning (FTL):} handles datasets
with limited overlap both in the sample space and in the feature spaces, as shown in Fig. \ref{fig:FTL}. Unlike VFL systems the number of common samples and thus labels that can be shared across the models is small thus limiting their reusability across the different models. A common transfer space is used for mapping the different feature spaces and exchanging knowledge.
Specifically, adopting transfer learning methods helps calculate feature values from distinct feature spaces, which is utilized for training local datasets \cite{sharma2019secure,kheddar2023deep}.

Let us consider a scenario with $N$ parties, where each party $i$, where $i \in \{1, 2, \dots, N\}$, has a local dataset denoted by $D_i$ and a task denoted by $T_i$. The goal in FTL is to jointly train a global model that leverages the knowledge learned from different tasks. The training process in FTL typically involves the following steps: (1) Each party $i$ performs local training on its own $D_i$ and task-specific model parameters $\theta_i$. During this step, the local model optimizes a task-specific loss function $L_i(\theta_i)$ to minimize the discrepancy between the local model's predictions and the corresponding labels in $D_i$. (2) The parties communicate and align their models (shared layers or parameters) to create a joint model architecture. (3) This transfer of knowledge, from party $i$ to the global model noted as $Ki(\theta)$, can occur through parameter sharing or feature extraction. steps (4) and (5) remain the same as HFL. The global loss function $L(\theta)$ in FTL typically includes a combination of task-specific loss terms and transfer-related terms. 
%
%Moreover, encryption algorithms, e.g., random masks are beneficial to ensure additional privacy preservation during gradient exchange between the server and clients \cite{so2021turbo}. 

In smart healthcare, FTL can support disease diagnosis by collaborating countries with multiple hospitals that
have different patients (sample space) and different monitor and therapeutic programs (feature space). In this way, FTL can enrich the shared AI model output for improving the accuracy of diagnosis.\\

% \begin{figure*}[ht!]
% \begin{center}
% \includegraphics[width=0.98\textwidth]{taxonomy_FL.eps}\\
% \end{center}
% \caption{Taxonomy of FL frameworks for CV applications.}
% \label{FedCV}
% \end{figure*}

\begin{figure*}[t!]
\centering
\includegraphics[scale=.69]{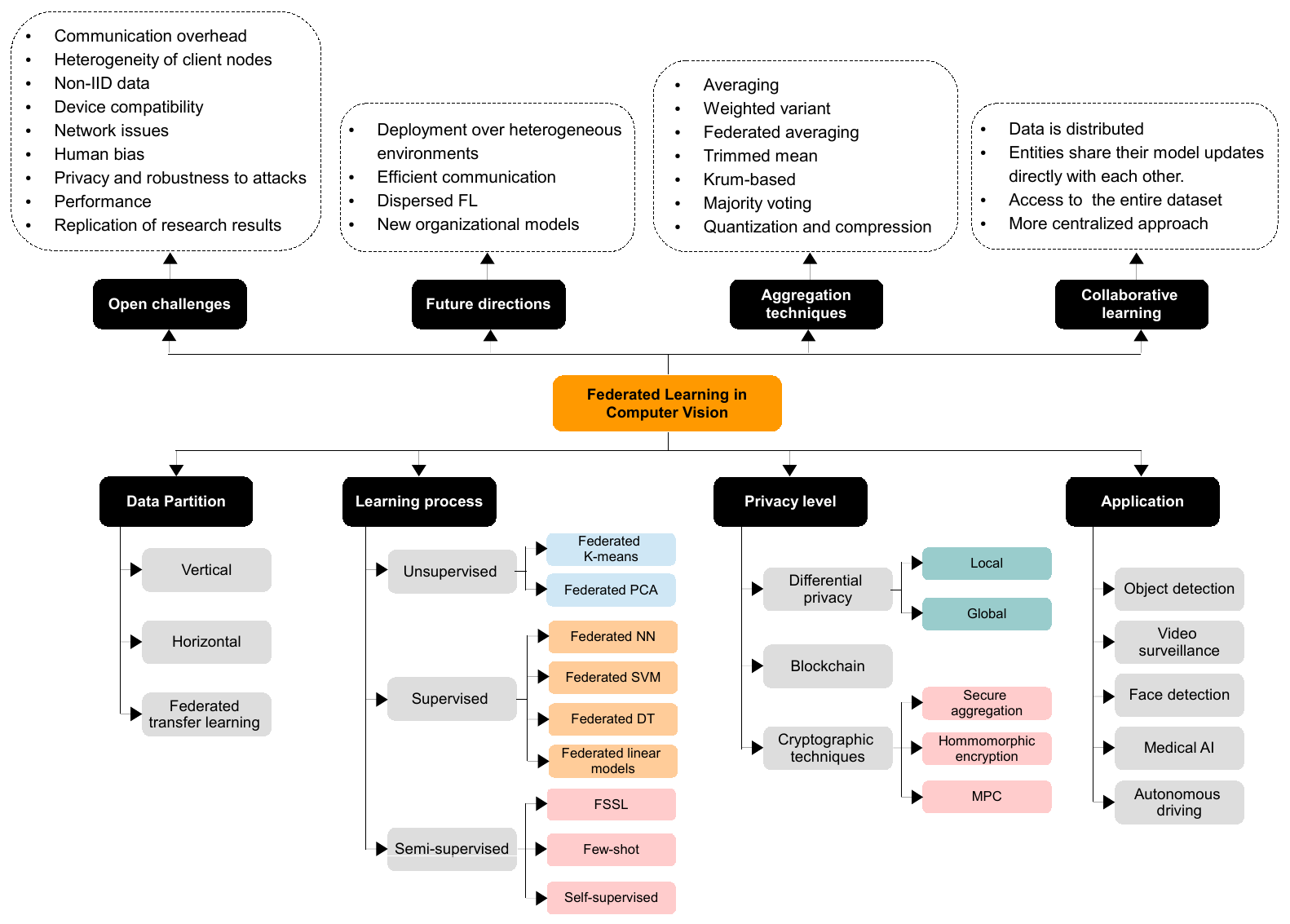}
\caption{Taxonomy of FL frameworks for CV applications.}
\label{FedCV}
\end{figure*}

\noindent\textbf{Collaborative learning (CL):} 
CL refers to a cooperative approach in which multiple entities actively participate and share their data and model updates to collectively train a shared model. It involves a collaborative effort where participants collaborate and contribute to the model's training by sharing their local knowledge and insights \cite{bjelobaba2023collaborative}. 
CL, within the context of FL, specifically refers to the scenario where participants actively collaborate and exchange data and model updates among themselves to collectively improve the shared model \cite{liu2022fedbcd}. In this sense, CL represents a specific instance of FL that emphasizes the collaborative nature of model training, where participants engage in direct data and model parameter sharing with each other, while still adhering to the principles of decentralized data ownership and privacy preservation characteristic of FL \cite{gabrielli2023survey}. CL can be justified as a form of FL based on the following key aspects:

\begin{itemize}
    \item \textbf{Data Distribution:} In CL, data is distributed across multiple entities or participants who collaborate to train a shared model. Similarly, FL involves training a model on decentralized data stored across multiple devices or servers, where each device contributes its local data.

   \item \textbf{Local Model Updates:} In both CL and FL, individual participants or devices perform local model updates using their respective data. These updates capture the specific patterns and characteristics present in their local data.

  \item \textbf{Aggregation and Collaboration:} Both approaches involve aggregating the local model updates from multiple participants or devices. In CL, participants typically share their model updates directly with each other, whereas in FL, the updates are aggregated on a central server or coordinator.

  \item \textbf{Privacy and Security:} CL and FL prioritize privacy and security. They aim to protect sensitive data by keeping it decentralized and only exchanging model updates or gradients instead of raw data. This ensures that participants' data remains private while enabling collaborative model training.

  \item \textbf{Iterative Improvement:} Both methods involve iterative improvement, where the shared model is updated based on the aggregated information from all participants. The process of local updates, aggregation, and model refinement is repeated until convergence is achieved.
\end{itemize}

\noindent While CL and FL share similarities, there are also some key differences between the two approaches:

\begin{itemize}
    \item \textbf{Data Ownership:} In CL, all participants typically have access to and ownership of the entire dataset. They actively share and collaborate on the data and model updates. In contrast, FL operates on decentralized data, where each participant retains ownership and control over their local data, which is not directly accessible to other participants.

   \item \textbf{Centralization vs. Decentralization:} CL tends to have a more centralized approach, where participants directly communicate and share their updates with each other. FL, on the other hand, employs a decentralized approach, where local updates are sent to a central server or coordinator for aggregation and model updates.

  \item \textbf{Communication Overhead:} In CL, participants typically need to establish direct communication channels to exchange data and model updates. This can require significant communication overhead, especially when the number of participants is large. FL reduces this communication overhead by relying on a central server or coordinator that facilitates the aggregation process.

  \item \textbf{Privacy and Security Focus:} While both approaches prioritize privacy and security, FL places a stronger emphasis on data privacy. FL minimizes the exposure of raw data by exchanging only model updates or gradients between participants and the central server. CL may involve more direct sharing of data or model parameters, which can introduce potential privacy risks.

  \item \textbf{Scalability:} FL is particularly suitable for large-scale distributed environments with a large number of participants or devices. Its decentralized nature allows for scalability and efficient training across a vast network of devices. CL may face challenges in terms of scalability, as direct communication and coordination among all participants become increasingly complex with a larger number of entities.

\end{itemize}

\noindent These differences highlight the varying degrees of decentralization, data ownership, communication, and scalability between CL and FL. While they share common principles, the specific implementation and focus of each approach can differ based on the context and requirements of the collaborative training scenario.

\subsection{Problem formulation}

Kone{\v{c}}n{\`y} et al.'s work has made FL well-known, but there are other definitions of the concept available in the literature \cite{konevcny2016federated}. FL can be realized through various topologies and compute plans despite a shared objective of combining knowledge from non-co-located data. This section aims to provide a detailed explanation of what FL is, while also highlighting the significant challenges and technical considerations that arise when using FL in CV. %in digital health.

%FL is a learning paradigm where multiple parties can collaboratively train without exchanging or centralizing datasets. A general FL formulation involves a global loss function, $\ell $, which is calculated from $K$ local losses, $\{\ell _{k}\}_{k=1}^{K}$. These local losses are computed from private data, $X_{k}$, which is stored on individual parties' systems without being shared among them. The global loss function is obtained by weighting the local losses.

The goal of FL is generally to build a global statistical model using data from many remote devices, which can range from tens to millions in number. This process involves minimizing an objective function that captures the desired characteristics of the model.

\begin{equation}
\underset{w}{\min } F(w),\text{ \ \ \ where \ \ \ }F(w):=%
\sum_{k=1}^{m}p_{k}F_{k}(w)
\end{equation}

\noindent where $m$ is the overall number of devices/centers, $F_{k}$ is the local objective function for the $k^{th}$ device/center, and $p_{k}$ represents relative impact of every device/center with $p_{k}\geq 0$ and $\sum\limits_{k=1}^{m}p_{k}=1.$

The empirical risk for a local dataset can be expressed using the local objective function $F_{k}$. The relative weight or influence of each device, represented by $p_{k}$, can be set by the user. There are two common choices for setting $p_{k}$: $p_k=\frac{1}{m}$ or $p_k=\frac{nk}{n}$, where $n$ is the total number of samples from all devices. 
Although this is a common FL objective, there are other alternatives, such as multi-task learning \cite{smith2017federated}, where related local models are learned simultaneously and each client corresponds to a task. Both multi-task learning and meta-learning allow for personalized or device-specific modeling, which can be a useful approach to handle the statistical heterogeneity of the data in FL. While FL and classical distributed learning both aim to minimize the empirical risk across distributed entities, there are several challenges that must be addressed when implementing this objective in a federated setting.

%%there exist other alternatives such as simultaneously learning distinct but related local models via multi-task learning \cite{smith2017federated} where each device corresponds to a task. Both the multi-task and meta-learning perspectives enable personalized or device-specific modeling, which can be a natural approach to handle the statistical heterogeneity of the data. At first sight, both FL and classical distributed learning share a similar goal of minimizing the empirical risk over distributed entities. However, implementing the above objective in a federated setting presents several fundamental challenges that must be addressed.

\subsection{Aggregation approaches in FL}

Following local training, the models are combined using an aggregation algorithm. To be more specific, each model takes an update step in its respective center $k$, utilizing a learning rate $\eta$ and the gradients $g_k$, so that \cite{linardos2021federated}:

\begin{equation}
    w_{t+1}^k= w_t-\eta.g_k, \forall k
\end{equation}

Subsequently, the weights are gathered into the global model in a manner that is relative to the sample size of each center \cite{linardos2021federated}: 

\begin{equation}
    W_{t+1}= \sum_{k=1}^m\frac{n_k}{n}w_{t+1}^k
\end{equation}

\noindent There are numerous FL aggregation techniques in the literature. In the following, we will discuss some of the most useful techniques specifically applicable to CV-based FL. These techniques include:\\

\subsubsection{Averaging aggregation} 
\begin{figure*}[t!]
\centering
\includegraphics[scale=.69]{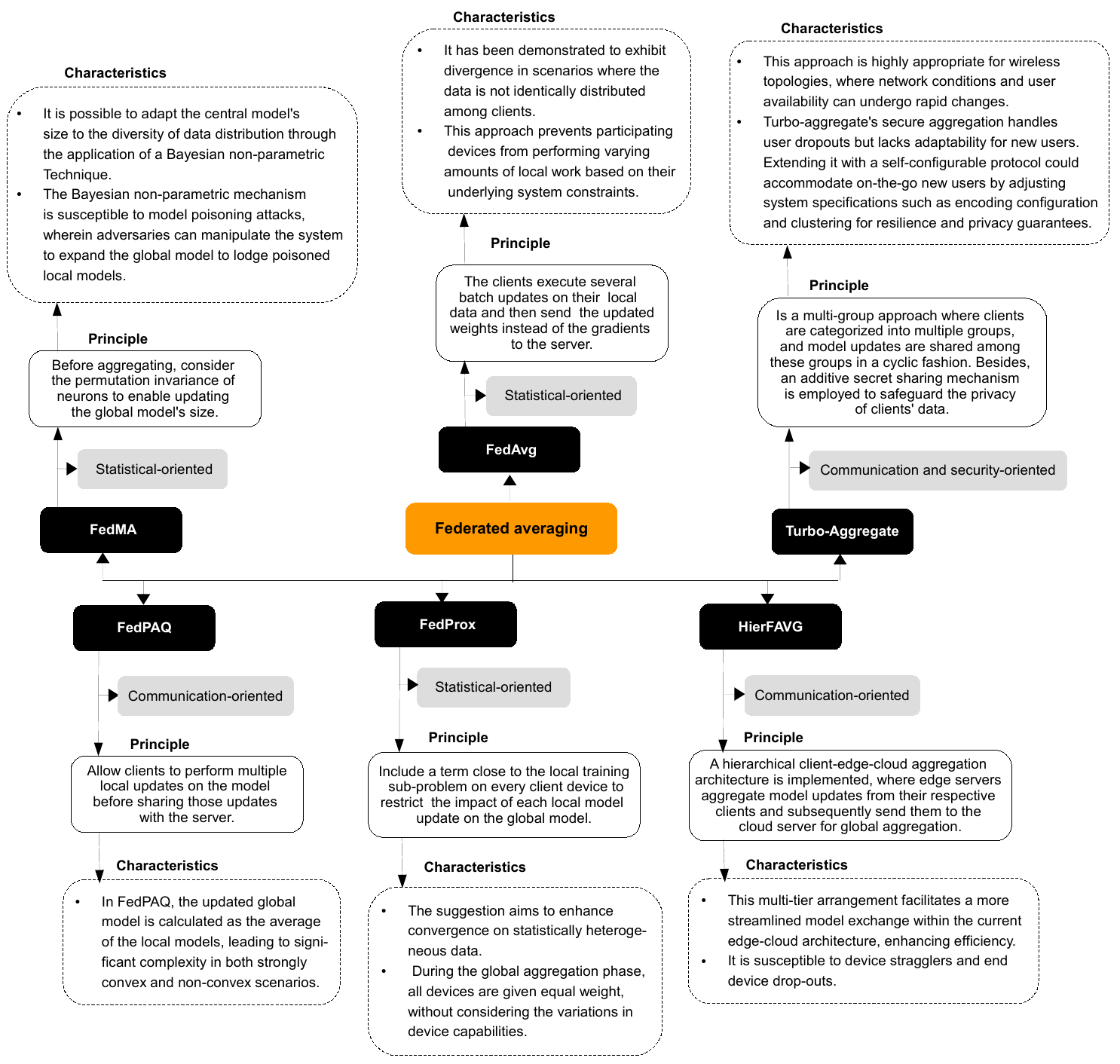}
\caption{Overview of key federated averaging Aggregation Approaches.}
\label{Fedavg}
\end{figure*}

There are many federated averaging (FedAvg) aggregation strategies in the literature. FedAvg algorithm is thoroughly detailed in \cite{wahab2021federated}. Fig. \ref{Fedavg} provides a summary of the main differences, characteristics, and principles among the alternatives.

\subsubsection{Progressive Fourier aggregation}
\begin{figure*}[t!]
\centering
\includegraphics[scale=1.4]{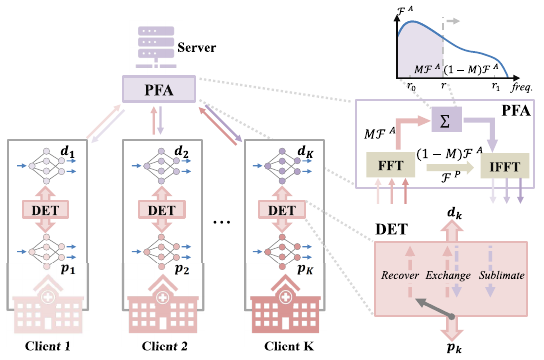}
\caption{Principle of PFA algorithm.  }
\label{PFA}
\end{figure*}

The paper \cite{chen2021personalized} addressed the retrogress and class imbalance problems using a personalized FL approach. Precisely, for better integrating the parameters of client models in the frequency domain, a progressive Fourier aggregation (PFA) is used at the server. Next, a deputy-enhanced transfer (DET) is designed at the clients' side to easily share overall knowledge with the personalized local model. In particular, the approach involves the development of PFA at the server, which ensures a stable and efficient gathering of global knowledge. This is achieved by gradually integrating client models from low-frequency to high-frequency. Additionally, the authors introduce a deputy model at the client's end to receive the aggregated server model. This facilitates the implementation of the DET strategy, which follows three types of decisions ($d_k$): Recover-Exchange-Sublimate. These steps aim to enhance the personalized local model by smoothly transferring global knowledge. In this process, the advantage of utilizing the Fast Fourier Transform (FFT) is to obtain both the amplitude map and the phase map of the parameters (conversion from parameters to map). The inverse FFT (IFFT) is used for the inverse operation (conversion from map to parameters). Fig. \ref{PFA} illustrates the principle of the PFA algorithm, which can be employed as an aggregation technique in FL. \\ 

\subsubsection{FedGKT aggregation}

\begin{figure*}[ht!]
\centering
\includegraphics[width=0.75\linewidth]{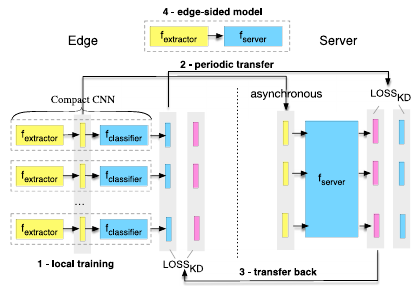}
\caption{Principle of FedGKT algorithm.}
\label{FedGKT}
\end{figure*}

FL group knowledge transfer (FedGKT) represents a streamlined FL approach tailored to edge devices with limited resources. Its primary objective is to combine the advantages of FedAvg and split learning (SL) by employing local stochastic gradient descent (SGD) training, similar to FedAvg, while simultaneously ensuring low computational burden at the edge, akin to SL. Notably, FedGKT facilitates the seamless transmission of knowledge from numerous compact edge-trained convolutional neural networks (CNNs) to a significantly larger CNN trained at a cloud server. Fig. \ref{FedGKT} illustrates the overview of FedGKT, (i) at the edge device, a compact CNN is employed, comprising a lightweight feature extractor and classifier, efficiently trained using its private data (local training). (ii) Following local training, consensus is reached among all edge nodes to generate uniform tensor dimensions as output from the feature extractor. Subsequently, the larger server model is trained, wherein the extracted features from the edge-side model serve as inputs. The training process employs a knowledge distillation (KD)-based loss function \cite{kheddar2023deep} to minimize the disparity between ground truth and soft label predictions, representing probabilistic estimations obtained from the edge-side model ( Periodic transfer). (iii) To enhance the edge model's performance, the server conveys its predicted soft labels to the edge, enabling the edge to conduct further training on its local dataset using a KD-based loss function with the server's soft labels ( Transfer back). As a result, both the server and edge models mutually benefit from knowledge exchange, leading to performance improvement. (iv) After completion of the training process, the final model is a fusion of the local feature extractor and the shared server model ( Edge-sided model) \cite{doshi2022federated}. The researchers who proposed FedGKT \cite{doshi2022federated}  claimed that it enables cost-effective edge training by significantly reducing computational expenses. In comparison to edge training with FedAvg, FedGKT demands only 9 to 17 times less computational power (measured in FLOPs) and requires a substantially smaller number of parameters, ranging from 54 to 105 times fewer.

\subsection{Privacy technologies of FL}
FL places a strong emphasis on privacy management, which involves analyzing security models in order to effectively protect personal information. In this section, various technologies that are currently used to safeguard privacy in the context of FL will be discussed.

% multi-party computation (SMC)
\subsubsection{Secure MPC model}
Secure MPC (SMPC) models involve data from multiple parties and use a well-defined simulation framework to provide safety certification. These models are designed to ensure that there is no interaction or sharing of knowledge data between the parties, meaning that users are unaware of the input and output data as well as any other information \cite{zhu2020relationship}. This zero-knowledge model can be considered a form of secure and complex computing protocol that is based on publicly available knowledge. Research has shown that SMPC can be used to create a security model that improves computing efficiency in low-security environments \cite{byrd2020differentially}.
The MPC protocol enables model training and verification without requiring users to share sensitive data. However, SMPC has certain limitations. It is unable to prevent servers from being curious and may not provide sufficient protection against privacy attacks from other clients. Additionally, the four-round interactive nature of the protocol can lead to wasted data and reduced model accuracy as the server does not have access to the client data until the submission phase is completed \cite{kanagavelu2020two}.

\subsubsection{Differential privacy}
Differential privacy technology, as indicated by past studies, has been employed as a protective measure for data privacy, where it processes and obscures sensitive attributes. This practice makes it challenging for third parties to identify individual users, thus rendering the data irrecoverable \cite{triastcyn2019federated,wu2022adaptive}. However, this method necessitates the transfer of data, which can negatively impact data accuracy, indicating a compromise between accuracy and privacy in such scenarios.
In addition, some researchers have suggested the incorporation of differential privacy within FL to safeguard client data by masking client contributions during training \cite{shah2021maintaining}. For instance, \cite{adnan2022federated} introduced a differentially private FL framework for analyzing histopathology images, which are considered some of the most intricate medical images. The authors assessed how data distribution, the number of healthcare providers, and individual dataset sizes influenced the framework's performance. They used the cancer genome Atlas (TCGA) dataset, an openly accessible repository, to mimic a distributed environment.
Furthermore, \cite{choudhury2019differential} presented an FL framework capable of generating a global model from distributed health data stored in diverse locations, without the need to move or share raw data. This framework also ensures privacy protection through differential privacy. An evaluation using real electronic health data from a million patients across two healthcare applications indicated the framework's success in delivering considerable privacy without compromising the global model's utility.

Differential privacy can be used in the training of DNN through the use of the differentially private stochastic gradient descent (DP-SGD) algorithm. \cite{ziller2021medical} introduces deepee, a free and open-source framework for differentially private DL that is compatible with the PyTorch DL framework for medical image analysis. This study uses parallel processing to calculate and modify the gradients for each sample. This process is made efficient through the use of a data structure that stores shared memory references to the neural network weights in order to save memory. It also provides specialized data loading procedures and privacy budget tracking based on the Gaussian differential privacy framework, as well as the ability to automatically modify the user-provided neural network architecture to ensure that it adheres to DP standards. Besides, \cite{zhang2022homomorphic} incorporates FL into the DL of medical image analysis models to enhance the protection of local models and prevent adversaries from inferring private medical data through attacks such as model reconstruction or model inversion. Cryptographic techniques such as masks and homomorphic encryption are utilized. Instead of relying on the size of the datasets, as is commonly done in DL, the contribution rate of each local model to the global model during each training epoch is determined based on the qualities of the datasets owned by the participating entities. Additionally, a dropout-tolerant approach for FL is presented in which the process is not interrupted if the number of online clients is above a predetermined threshold.

%We compared the performance of private, distributed training with conventional training and found that it can achieve similar results while providing strong privacy protections.

\subsubsection{Homomorphic encryption}
Homomorphic encryption is a type of encryption used in the ML process that uses parameter exchange to protect the privacy of user data. Unlike differential privacy, it does not transmit the data or models themselves and encrypts the data without allowing it to be discovered. This makes it less likely for the original data to be leaked. The additive homomorphic encryption model is the most commonly used version in practice.
Besides, despite the benefits of FL, there is a risk that private or sensitive personal data may be exposed through membership attacks when model parameters or summary statistics are shared with a central site. To address this issue, \cite{stripelis2021secure} presents a secure FL-based framework that employs fully homomorphic encryption (FHE). In doing so,  the Cheon-Kim-Kim-Song (CKKS) construction has been used,
which enables the execution of approximate calculations, on real and floating-point numbers benefiting from ciphertext rescaling and packing. Moving on, \cite{kumar2022blockchain} introduces a homomorphic encryption and blockchain-based privacy-preserving aggregation framework for medical image analysis.
This allows hospitals to collaborate and train encrypted federated models while maintaining data privacy, using blockchain ledger technology to decentralize the FL process without the need for a central server.
Specifically, homomorphic encryption protects the privacy of the model's gradients. This framework involves training a local model using a capsule network for the segmentation and classification of COVID-19 images, securing the local model with the homomorphic encryption scheme, and sharing the model over a decentralized platform using a blockchain-based FL algorithm.

%\section{Overview of FL frameworks}
\subsection{Learning process}
\subsubsection{Supervised FL}
\begin{figure}[t!]
\centering
\includegraphics[width=0.7\linewidth]{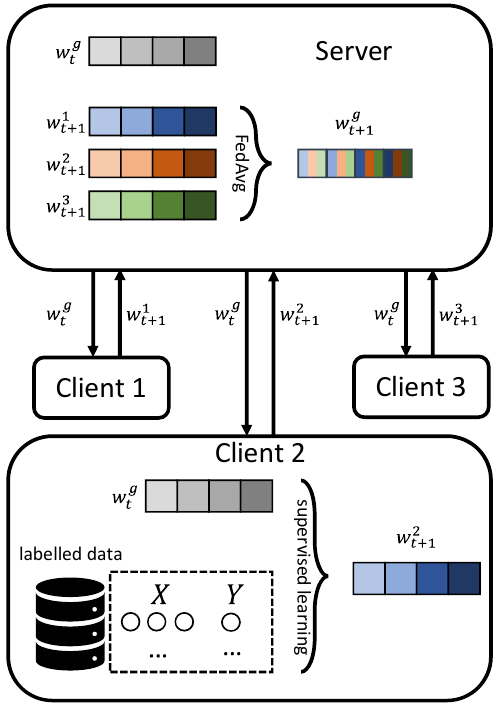}
\caption{The basic setup of a typical FL system with supervised learning involves several steps. Initially, the server picks three clients and transmits the global model $w_t^g$  to them. Subsequently, these clients employ their labeled data to update the global model $w_t^g$, generating their respective local models. Once this process is completed, the clients send their updated local models back to the server. The server then aggregates these local models into a new global model using the FedAvg algorithm \cite{zhao2020semi}.}
\label{flsl}
\end{figure}

Supervised FL trains ML models on sensitive labeled data across multiple devices and learns to predict an output based on the input data. Each device runs the model and updates its parameters based on its local data, and then the updates are aggregated to improve the global model. The supervision is ensured through the presence of labeled data in the participating entities' local datasets, and the learning process can be implemented using support vector machines (SVM) \cite{brisimi2018federated}, linear models (LMs), neural networks (NN), or decision trees (DT) \cite{li2021survey}. Fig. \ref{flsl} illustrates an example of a typical supervised FL with three clients.

\subsubsection{Unsupervised FL}
Unsupervised FL is a variant of FL in which the participating devices do not have access to labeled data. Instead, they must rely on unsupervised learning techniques, such as clustering or dimensionality reduction, to learn useful representations of the data. This can be useful in scenarios where collecting and annotating labeled data is difficult or expensive, or where the data is highly sensitive and cannot be shared with a centralized server. The learning process can be implemented using  federated principal component analysis (PCA) \cite{grammenos2020federated}, federated k-means \cite{kumar2020federated}, etc.

\subsubsection{Semi-supervised FL}
\begin{figure*}[t!]
\centering
\includegraphics[width=0.9\linewidth]{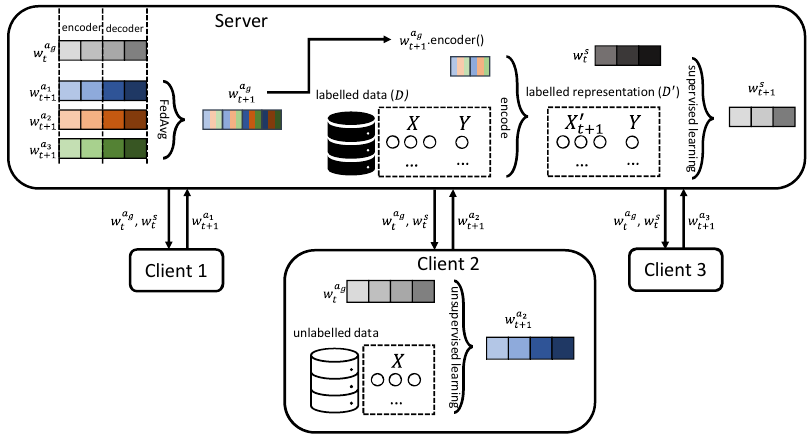}
\caption{The structure of our semi-supervised FL system involves a server and three clients, which adhere to the standard FL process to update a global autoencoder $w_{t}^{a_g}$. Next, the server employs the encoder from $w_{t+1}^{a_g}$ to encode the labeled dataset $D$ on the server, resulting in a labeled representation dataset $D'$. Afterward, the server utilizes supervised learning with $D'$ to train a classifier $w_{t}^{s}$, producing a new classifier $w_{t+1}^{s}$ \cite{zhao2020semi}.}
\label{ssfl}
\end{figure*}

FL allows training ML algorithms with a semi-supervised learning (SSL) process on remote datasets without the need to share the data itself. However, data annotation remains a challenge, especially in fields like medicine and surgery where specialized knowledge is often required. To address this issue, semi-supervised FL has recently been used where the participating devices have access to a dataset that contains both labeled and unlabeled examples. The labeled examples are used to train a model using supervised learning techniques, while the unlabeled examples are used to learn additional useful features using unsupervised techniques such as clustering or dimensionality reduction. Fig. \ref{ssfl} illustrates an example of a typical semi-supervised FL with three clients. 

In this direction, Kassem et al. \cite{kassem2022federated} propose FedCy, a federated SSL (FSSL) system that combines FL with self-supervised learning to improve the performance of surgical phase recognition using a decentralized dataset containing both labeled and unlabeled videos. FedCy uses temporal patterns in the labeled data to guide the unsupervised training of unlabeled data towards task-specific features for phase recognition. This scheme outperforms state-of-the-art (SOTA) FSSL methods on the task of automatically recognizing surgical phases using a multi-institutional dataset of laparoscopic cholecystectomy videos. Additionally, it learns more generalizable features when tested on data from an unseen domain.
\cite{rehman2022federated} examines the effectiveness of SOTA video SSL techniques when used in a large-scale FL setting, as simulated using the kinetics-400 dataset. The limitations of these techniques in this context are identified before introducing a new federated SSL framework for a video called FedVSSL. FedVSSL incorporates various aggregation strategies and partial weight updates and has been shown through experiments to outperform centralized SOTA by 6.66\% on UCF-101 and 5.13\% on HMDB-51 in downstream retrieval tasks. 

Moving forward, \cite{dave2022spact} presents a self-supervised privacy preservation framework for action recognition, namely SPAct. It includes three main components: (i) an anonymization function, (ii) a self-supervised privacy removal module, and (iii) an action recognition module. A minimax optimization strategy is used to train this framework, which minimizes the action recognition cost function and maximizes the privacy cost function through a contrastive self-supervised loss. By using existing protocols for known action and privacy attributes, this framework achieves a good balance between action recognition and privacy protection, similar to the current SOTA supervised methods. Additionally, a new protocol to test the generalization of the learned anonymization function to novel action and privacy attributes is introduced.

In \cite{li2022fedutn}, FedUTN is proposed which is an FL allowing each client to train a model that works well on both independently and identically distributed (IID) and non-independent and identically distributed (non-IID) data. In this framework, each party has two networks, a target network, and an online network. FedUTN uses the online network parameters from each terminal to update the target network, which is different from the method used in previous studies. FedUTN also introduces a new control algorithm for training. After testing, it was found that FedUTN's method of aggregation is simpler, more effective, and more robust than other FL algorithms, and outperforms the SOTA algorithm by 0.5\%-1.6\% under normal conditions.
In \cite{shome2021fedaffect}, a few-shot FL framework utilizing a small amount of labeled private facial expression data to train local models on individual devices is proposed. They are then aggregated in a central server to create a globally optimal model. FL is also used to update the feature extractor network on unlabeled private facial data from user devices to learn robust face representations. 

\subsection{Evaluation of FL-based CV scheme}

To assess the efficiency of FL systems in CV, it is essential to employ metrics that are responsive in gauging the performance of implementing FL across various CV applications. Besides the famous metrics that are commonly employed in ML and DL, such as accuracy, F1-score, and area under the curve (AUC) \cite{kheddar2023deepSteg},  other metrics are summarized in Table \ref{metrics}.

\begin{table*}[t!]
\caption{ The evaluation metrics used in the suggested FL-based CV schemes.}
\label{metrics}

\renewcommand{\arraystretch}{1.6} % Reset to original spacing
\begin{tabular}{lcm{100mm}}
\hline
Metric & Math description & Characteristic \\ 
\hline 
AP & \(\displaystyle \frac{1}{N} \sum_{k=1}^{N} P(k) \cdot \text{{rel}}(k)\) &  The average precision (AP) is commonly used in information retrieval and object detection tasks. It measures the precision of a model across different levels of recall. \(N\) represents the total number of retrieved items, \(P(k)\) denotes the precision at the \(k\)-th retrieved item, and \(\text{{rel}}(k)\) indicates whether the \(k\)-th retrieved item is relevant (1 if relevant, 0 if not relevant).
\\ [0.5cm]
mAP & \(\displaystyle \frac{1}{N_{\text{{class}}}} \sum_{c=1}^{N_{\text{{class}}}} \text{{AP}}_c \)
 & The mean average precision (mAP) is often employed to assess the performance of models that generate multiple predictions or retrieve ranked results. \(N_{\text{{class}}}\) represents the total number of classes, \(\text{{AP}}_c\) denotes the AP for class \(c\).\\[0.5cm]
MAE & \(\displaystyle \frac{1}{H\times W}\sum_i^H \sum_j^W (P(i,j)-G(i,j))  \) & The mean absolute error (MAE) measures the average absolute difference between each pixel in a normalized prediction saliency map $P(i,j)$ and its corresponding pixel in the binary ground truth mask $G(i,j)$. $H$, and $W$ represent the image's height and width.\\[0.5cm]

RMSE & \(\displaystyle \sqrt{\frac{1}{{M \times N}} \sum_{i=1}^{M} \sum_{j=1}^{N} (P(i,j) - G(i,j))^2}
\) & Is a widely used metric to measure the difference between prediction and ground truth. It calculates the average magnitude of squared differences between corresponding elements in matrices, providing a measure of accuracy.\\
\hline
\end{tabular}
\end{table*}

\section{Applications of FL in CV}
\label{applications}

FL has diverse applications. It enables privacy-preserving object detection, face detection, and video surveillance in smart environments. It also facilitates advancements in healthcare and medical AI, and plays a crucial role in autonomous driving within CV. Below, the survey provides detailed information about each of these domains.

\subsection{Object detection }

A special CV task is the segmentation of images and more specifically the detection of salient objects in one or multiple images. Salient object detection (SOD) is an important pre-processing step in many CV tasks, such as object detection and tracking, semantic segmentation, and the interaction between robots and humans (e.g. for tracing human hands during imitation learning scenarios). 

\subsubsection{CL for object detection}
CL techniques that combine the merits of multiple feature learning techniques from images seem to be the most promising technique. The main issue to handle is the separation of the object from its background, any foreground occlusions, or noise. Other important issues related to the size and variety of objects to be detected in the same image and the need to track objects across consecutive or groups of images. For example, the work in \cite{ji2020accurate} emphasized on the use of RGB-D methods that combine RGB images with depth images in order to improve the SOD across multiple images. They proposed a CL framework (CoNet), which combines the edge information extracted from low-level features of the images with the merits of a spatial attention map that detects salient features in the images and depth images that better locate salient objects in a scene. The three different collaborators are combined in a knowledge collector module that first concatenates salient and edge features to jointly learn the boundaries and locations of salient objects and then uses the depth information to separate salient regions from their background. Another work \cite{fan2021group} presents a NN architecture for content-aware segmentation of sets of images that employs co-saliency maps generated from the input images using a group CL framework (GCoNet). The proposed method outperforms standard SOD alternatives and is capable of detecting co-salient objects in real-time. The main criteria employed are the compactness of the extracted objects within the group of images and the separability of objects from other noise objects in the scenes. 

Moving on, the research work on hand detection and tracking over long videos \cite{teeparthi2021fast, teeparthi2021long} focused on the need for fast and long-term object detection even in the presence of temporal occlusions and changes among frames. In order to tackle these problems, the researchers suggested the use of a hand detection model (Based on faster R-CNN) combined with the projection across an image (frame) sequences in order to detect clusters of bounding boxes that remain almost stable across frames, even if they are occluded in some of the frames. In order to handle the problem of occlusions, aspect changes and articulations that may hinder the proper detection of objects in uncontrolled scenes, \cite{huang2010high} propose a method that first detects object parts and consequently tries to associate them and detect the object. The detected object parts are called granules, i.e. small areas with simple properties (e.g. color) that separate them from their context. Using a combinatorial optimization method, based on simulated annealing, they learn how to associate the proper neighboring granules that compose the object. The problem of overlapping objects is also discussed by \cite{fang2021collaborative}, where the authors propose a new object detection algorithm that improves the localization of the detected objects and suppresses redundant detection boxes, using a better loss function. 

Moreover, CL for medical image segmentation and classification has been applied by \cite{zhou2019collaborative}. The authors distinguish between the image segmentation and annotation of segments task, and that of detecting the severity of the disease by considering the image as a whole. They propose the use of CL method for disease severity classification that is based on attention over the detected and annotated segments. A weakly supervised framework for CL is proposed by \cite{wang2018collaborative} for allowing object detection using labels at the image level. The proposed framework combines a weakly supervised learner (i.e. a two-stream CNN built on VGG16) with a strongly supervised learner (i.e. faster-RCNN) and trains the two subnetworks in a collaborative manner. The former subnetwork optimizes the multi-label classification task, whereas the later optimize the prediction consistency of the detected objects' bounding boxes. Another weakly supervised object detection approach that employs image-level labels in order to learn to detect the accurate location of objects in the training images has been presented by \cite{zhang2019leveraging}. A CL framework has first been trained on the image-level labels aiming to optimize the image-level classifier and to assign higher weights to instances with more confidence (i.e. without much noise, with fewer objects and simpler backgrounds). Then, a curriculum learning component is employed for guiding the learning procedure of object localization and classification. In \cite{chen2020fcc}, the authors also propose an object detection method from remote sensing images, which combines a weakly supervised detector for image-level labels and a strongly supervised detector for object localization. In a similar manner, \cite{liang2021semantic} split the salient object detection task into two sub-tasks that are examined in parallel. The first task relates to the estimation of internal semantics and the second to the prediction of the object boundaries. VGG-16 is used as a basis for feature extraction from images and an additional layer is used to define their semantics. Two decoders are combined for the second part. The first detects the broader area of interest in the images and the second performs a fine-grained boundary detection. The CL network joins the two networks in order to efficiently fuse semantics and boundaries and extract the saliency maps of each image.  Combining an image enhancement network with the object recognition network in order to be able to recognize objects in images of extremely low resolution is proposed in \cite{seo2019object}. The two networks were trained using a CL technique in which the knowledge from the object recognition network is used to enhance the low-resolution images that are given as input to the other network. The enhanced images are then fed to the classifier to improve its performance. Four different losses (reconstruction, perpetual, classification and edge loss) are combined, in order to optimize the object recognition performance. Transferring the knowledge distillation method in a distributed and CL setup has been proven beneficial for online object detection tasks \cite{guo2020online}. The student models (one in each node) are trained separately using as their teacher model, an ensemble that comes from the fusion of the student logits. The teacher's knowledge is then distilled back to the students in the form of a soft target. This logic can easily be ported to the FL paradigm as an alternative to FL-based averaging (FedAvg) or more proficient averaging techniques.

\subsubsection{FL for object detection}
The learning process for object detection is usually handled as a centralized offline task, but in resource-restricted environments and applications that need mobility, privacy, and security decentralized and distributed approaches, as well as cloud-edge collaborative approaches have been proposed. In an attempt to optimize object detection performance, authors in \cite{yu2019federated} proposed an FL approach that improves the performance of the federated averaging model aggregation over independent and identically distributed (IID) data. The main claim of their work is that SOTA DL CNN models have been trained in controlled, centralized and balanced datasets and cannot perform well on non-independent and identically distributed (non-IID) data. Since in cases that need privacy and security, such as in medical images for example, the data can be non-IID, the use of simple federated algorithms, such as FedAvg, can be problematic. For this purpose, they propose a weighted variant of FedAvg that improves FL performance in non-IID image data.  Using collaborative intelligence on the edge, or balancing between the edge and the cloud is an efficient FL paradigm that can improve the performance of object detection tasks. The work of \cite{choi2018deep} has shown that splitting the deep NN models between the cloud and the edge, apart from being privacy-preserving and secure can also be more efficient for object detection tasks. Efficiency can be achieved by quantizing and compressing the tensors of the first layers before sending them to the final layers that reside in the cloud. Depending on where the split is performed, different compression techniques (lossless or lossy ones) are preferred. The same network split strategy (splitNN) has been used in \cite{vepakomma2018split} for protecting the privacy of medical data. More specifically, the client nodes train the network up to a certain layer and the outputs are sent to the server to perform the rest of the training. The inverse process takes place during the back-propagation of the gradients. In order to further protect the labels of the training samples from exposure to the server, the authors also propose a u-shaped forward and backward propagation process in which the first and last layers of the network (along with the training labels) are kept in the clients. Finally, they propose a vertically partitioned data model in which multiple clients train the same first layers of their networks in parallel and send the outputs to the server that concatenates them before proceeding with the remaining layers. This way the clients share the model (at least the last layers) without sharing their data.

\subsection{Self-supervised-based FL}

Self-supervised learning (SSL) and its variants, momentum contrast (MoCo), bootstrap your own latent (BYOL), and simple siamese (SimSiam), are powerful techniques for learning representations from centralized data \cite{he2022hybrid}. FL has been combined with SSL to address privacy concerns with decentralized data. However, there is a lack of in-depth understanding of the fundamental building blocks for federated self-supervised learning (FedSSL).  The reference \cite{he2022hybrid} introduces a federated hybrid self-supervised learning (FedHSSL) framework that combines VFL and SSL techniques such as SimSiam, addressing data deficiency. FedHSSL utilizes cross-party views and local views of aligned and unaligned samples within each party to enhance representation learning. It also incorporates partial model aggregation using shared generic features among parties. Experimental results show that FedHSSL outperforms baseline methods, particularly when labeled samples are limited. The researchers also analyze privacy protection mechanisms in FedHSSL, demonstrating its effectiveness in thwarting label inference attacks. Moving forward, Zhuang et al. \cite{zhuang2022divergence} propose a generalized FedSSL framework that accommodates existing SSL methods based on Siamese networks and allows for flexibility in future methods. Through empirical study, the authors uncover insights that challenge the conventional wisdom, showing that stop-gradient operation is not always necessary in FedSSL and retaining local knowledge of clients is beneficial for non-IID data. Building on these insights, the authors propose a new model update approach called federated divergence-aware exponential moving average (FedEMA) update, which adaptively updates local models of clients using exponential moving average of the global model with dynamically measured decay rate based on model divergence. Experimental results demonstrate that FedEMA outperforms existing methods on linear evaluation. The authors hope that this work provides valuable insights for future research in the field of FedSSL. Saeed et al. \cite{saeed2020federated} propose a self-supervised approach called scalogram-signal correspondence learning based on wavelet transform (WT) to learn representations from unlabeled sensor inputs such as electroencephalography, blood volume pulse, accelerometer, and WiFi channel-state information. The proposed method addresses issues related to privacy, bandwidth limitations, and cost of annotations associated with centralized data repositories for supervised learning. The auxiliary task involves a deep temporal neural network to determine if a given pair of a signal and its complementary view align with each other, optimized through a contrastive objective. The learned features are extensively evaluated on various public datasets, demonstrating strong performance in multiple domains. The proposed methodology achieves competitive performance with fully supervised networks and outperforms autoencoders in both central and federated contexts. Notably, it improves generalization in a semi-supervised setting by reducing the volume of labeled data required through leveraging self-supervised learning.

Yan et al.~\cite{yan2023label} propose a robust and label-efficient self-supervised FL (LE-SSFL) framework for medical image analysis. The method utilizes a Transformer-based self-supervised pre-training approach on decentralized target task datasets using masked image modeling. This enables robust representation learning and effective knowledge transfer. Experimental results on simulated and real-world medical imaging non-IID federated datasets demonstrate improved model robustness against data heterogeneity. The approach achieves significant enhancements in test accuracy on retinal, dermatology, and chest X-ray classification without additional pre-training data, outperforming a supervised baseline with ImageNet pre-training. The FedSSL pre-training approach demonstrates better generalization to out-of-distribution data and improved performance with limited labeled data compared to existing FL algorithms. Table \ref{tab:obj} presents of summary of FL frameworks proposed for object detection tasks.

\begin{table*}[t!]
\small
 \caption{A summary of FL frameworks proposed for object detection tasks. Learning type (LT), best performance (BP), project link availability (PLA).} \label{tab:obj}

\begin{tabular}{p{0.3cm}p{0.3cm}p{1.2cm}p{2cm}p{4.5cm}p{1.5cm}p{4cm}p{1cm}}
  
\toprule
Ref. &LT & Model & Dataset & Description & BP & Limitations & PLA   \\ \hline
   
\cite{fang2021collaborative} & CL & ResNet50, YOLO 3 &  Pascal Voc 2007, MS COCO & A new effective loss function handling both  overlapping and non-overlapping boxes & AP=17.08 & only applicable to one stage detectors & Yes\tablefootnote{\url{https://github.com/JaryHuang/awesome_SSD_FPN_GIoU}} \\

\cite{wang2018collaborative} & CL & WSDNN, Faster R-CNN & Pascal 2007, 2012 & An end-to-end to end weakly supervised collaborative learning framework to improve model accuracy & AP=20.3 &  Achieves maximum performance with a high number of iterations. & No\\

\cite{zhang2019leveraging} & CL & VGG16, Fast RCNN & Pascal Voc 2007, test-2007, test-2010, COCO 2014 & A novel collaborative self-paced curriculum learning exploiting prior knowledge to enhance the accuracy of weakly object detection &  mAP=7.2 & Generalizability needs to be verified with other tasks and datasets. & No\\

\cite{chen2020fcc} & CL & FCCNET & TGRS-HRRSD, DIOR & A new CNN model based on several mechanisms that is trained by alternating strongly and weakly detectors & mAP=0.2 & The results drastically between classes. &  No\\

\cite{liang2021semantic} & CL & SDCLNET, VGG, ResNet & ECSSD, Pascal-C, DUTS, HKU-IS, DUT-OMRON, SOD & A new model leveraging collaborative learning to optimize to sub goals, internal semantic estimation and boundary detail prediction, to improve the accuracy of saliency maps &  MAE=0.03 & Considers only the accuracy while disregarding aspects such as network space redundancy. &  No\\

\cite{seo2019object} & CL & ResNet-152 & CIFAR-10, CIFAR-100, ImageNet & improvement of object detection in low-resolution images through use  of two neural networks trained jointly & ACC=7.31 &  Acceptable yet unsatisfactory results. &  No\\

\cite{guo2020online} & CL & ResNet & MS COCO, CIFAR, ImageNet & A new online knowledge dist distillation method leveraging several networks trained in parallel acting all as students & ACC=72.9 & Generalization issues with limited data. & No \\ 

\cite{yu2019federated} & FL & SSD300 & Pascal Voc 2007  & FL is used to train the model with a large number of features. KLD measures weights divergence between non-IID models. The scheme reduces the impact of abnormal weights in FedAvg using SSD. & FedAvg=76.6 (NonIID) & Generalizability needs to be verified with other datasets. & No\\

\bottomrule
\end{tabular}
\end{table*}

\subsection{Face detection}
The detection of face mask wearing is another CV task that can be significantly benefited by FL \cite{zhu2021masked}. When visual data are collected by surveillance cameras, the monitoring of face mask-wearing can be split into two separate tasks: i) the detection of faces, ii) the detection of proper mask-wearing. The second task poses a risk to user privacy, making FL the preferred approach. The proposed scheme utilizes a dilation retina-Net face location network for \textit{detecting} faces, including both clear and occluded ones, in a dense crowd. Additionally, an SRNet20 network has been employed to \textit{process} the detected faces. The second network is trained using multiple client nodes that share their models with a central server, which aggregates the client models.

Furthermore, a related-face detection task is facial expression recognition, which has numerous applications in the field of affective computing. The study conducted by \cite{shome2021fedaffect} proposes a few-shot meta-learning-based FL framework that requires only a limited number of labeled images for training. The local devices process numerous unlabeled face images, which are utilized to train an encoder network for learning face representations. Periodically, a central server aggregates the models using the FedAvg algorithm to enhance representation learning. Additionally, a second few-shot learner employs a small number of labeled private facial expressions to train local models, which are subsequently aggregated in the central server.

\subsection{Video surveillance and smart environments}

An application that attracts the interest of CV researchers in terms of balancing the workload between centralized (cloud) and distributed (edge-based) architectures is video surveillance~\cite{himeur2023video}. Edge computing has offered several solutions in the direction of workload balancing. However, it added new challenges concerning the compression and filtering of data that is transmitted over the communication network, and the fragmentation of knowledge across the edge devices. In order to tackle the aforementioned projects,  \cite{sada2019distributed} propose a multi-layer design in order to minimize the communication between the edge and the cloud and employ FL to update the detection models without sending training data to the cloud. The multi-layer architecture comprises a data acquisition layer on the edge that performs compression and filtering, an object detection and identification layer with edge servers that are specialized in specific tasks (e.g. face recognition, licence plate detection, etc.) and a FL layer that solves tasks on the edge and marks data and models that must communicated to the cloud. The information transferred to the cloud usually comprises model parameters and the respective weights.

A generic FL architecture that processes sensor data for human activity recognition has been presented by \cite{concone2022federated}. The proposed FL architecture is based on a federated aggregator that is trained using local data on the edge nodes. The aggregator sends the models to a central entity where the Federated Average algorithm is employed to generate a common model that is distributed back to the edge nodes. The federated aggregator in the video surveillance scenario could be a local server that is trained on the edge using local data.
% The following document is in Chinese. Better not add it to the references    \cite{yu2020federated} 
Moving on, \cite{liu2020federated} introduces a FL framework based on a lightweight version of the Dense-MobileNet model for processing aerial images collected by unmanned aerial vehicles (UAV) swarms. The UAV swarms used for vision sensing collect haze images, and are supported by ground sensors that collect information about air quality. Each UAV employs a ML model in order to correlate air quality with the haze images and shares its trained gradient with a central server. The server combines the gradients a learns a global model, which is then used to predict air-quality distribution in the region. 
Besides, platforms that supports FL powered applications of CV in \cite{liu2020fedvision,catalfamo2022platform}.  Lia et al. in \cite{liu2020fedvision}  demonstrate FedVision in a fire detection task, using a YOLOv3 model for object detection and they report its use on three pilots concerning safety hazard detection, photovoltaic panel monitoring and suspicious transaction monitoring in ATMs via cameras. Catalfamo et al. \cite{catalfamo2022platform}  thoroughly examine the use of edge federation for implementing ML-based solutions. The platform is introduced for deploying and managing complex services on the edge that utilizes micro-services, which are small, independent, and loosely-connected services. This platform enables the management of these services across a network of edge devices by abstracting the physical devices they run on. The the effectiveness of this solution was demonstrated through a case study involving video analysis in the field of morphology.

\subsubsection{Action recognition}

Using knowledge distillation, \cite{jain2021federated} allows client nodes with limited computational resources to execute action recognition by performing model compression at the central server on a large-scale data repository. 
In doing so, fine-tuning is used since small datasets are not appropriate for action recognition models to learn complex spatio-temporal features. 
Moving forward, an asynchronous federated optimization is adopted and the convergence bound has been shown since the present clients' computing resources were heterogeneous. In a different paradigm, a driver action recognition system is built by \cite{zhang2021driver} using FL. The latter has been used for model training to protect users' privacy while enabling online model upgrades. Similarly, in \cite{doshi2022federated}, an FL-based driver activity recognition system is implemented by training the detection model in a decentralized fashion.
FedAvg and FedGKT have been implemented and their performance has been demonstrated on the 2022 AI City Challenge. Zhao et al. \cite{zhao2020semi} proposed a semi-supervised FL for activity recognition, which helps edge devices conducting unsupervised learning of general representations using autoencoders and non annotated local data. In this case the cloud server performed supervised learning by training activity classifiers on the learned representations, and annotated data.

% \cite{} 

% \cite{} 

% \cite{} 

% \cite{} 

% \cite{} 

% \cite{} 

% \cite{} 

% \cite{} 

% \cite{} 

% \cite{} 

% \cite{} 

\subsubsection{Crowd counting}

When more refined tasks are assigned to CV algorithms in dense crowds, such as the detection of face mask wearing \cite{zhu2021masked}, the difficulty increases since the target objects (i.e. masks) are in different scales and occlusions. FL is a promising solution that enhances the privacy of individuals. However, scale variations, occlusions and the diversity in crowd distribution are still the open issues that demand efficient detection techniques, such as deep negative correlation learning \cite{shi2018crowd}, relational attention \cite{zhang2019relational}, etc.

Crowd counting is a complex CV task especially when multiple sensors (i.e. cameras) are combined. FL approaches can be helpful since they allow the distributed trainers to exchange their models and improve their performance quickly, especially when a centralized trainer is able to periodically validate and improve the quality of the aggregated model \cite{giorgas2020online}. However, when the distributed nodes are not trusted in advance, or when the quality of their data is ambiguous, more control mechanisms and incentives are needed to avoid the deterioration of the resulting model.
When the distributed trainers have to co-operate in order to reach a consensus, either it is for crowd counting or for any other task, it is important to provide them enough incentives in order to become trustful \cite{jiang2020federated}. Blockchain-based approaches \cite{bao2019flchain} manage to distribute the incentive among the trained models and evaluate their reliability in order to fairly partition any potential profit (or trust). This way, they allow FL algorithms to become more robust, by including trustful trainers and honest reporters that detect misbehaviors and block them from the FL process.

\subsubsection{Anomaly detection}
An interesting data mining task with many CV applications is anomaly detection. It involves the identification of strange patterns in data, which may indicate a fake or incorrect situation. Several cutting-edge ML and DL algorithms have been developed in the literature in order to detect and prevent such incidents. When it comes to CV various DL architectures, from variational autoencoders (VAE), generative adversarial networks (GANs) or recurrent neural networks (RNNs) can be trained to detect anomalies in video sequences \cite{kiran2018overview}. 
The capitalization of the use of autoencoders and FL for detecting anomalies is introduced in  \cite{singh2021anomaly}. The bipartite structure of autoencoders, and their ability to reconstruct the input in their output, allows the detection of anomalies based on the errors spotted during the regeneration of the input. The authors train two clients using different parts of the training dataset and they merge the resulting models using FL library called PySyft for secure and private DL. They employ the MSE loss metric to compare the input and output and examine whether it exceeds a threshold in order to decide whether it is an anomaly or not. Bharti et al. \cite{bharti2022edge} have proposed an edge-enabled FL approach for the automatic inspection of products using CV. Their main basis is SqueezeNet, a lightweight model pre-trained on the ImageNet dataset, which is able to identify 100 different types of objects. Although the image processing model has been trained using multiple images of normal products the various defects are detected using an anomaly detection algorithm. SqueezeNet acts as a feature extractor from images, which are then fed to a dense layer with as many output neurons as possible anomaly classes, plus one for the normal products. In its federated version, the edge server aggregates the various local models and computes a new global model that is reshared with the local nodes. Table \ref{tab:FL} summarizes some of the FL frameworks proposed for video surveillance and smart environments.

\begin{table*}[t!]
\caption{Summary of FL frameworks proposed for video surveillance and smart environments. Best performance (BP), project link availability (PLA).} \label{tab:FL}

\renewcommand{\arraystretch}{1.6} % Reset to original spacing

\begin{tabular}{p{0.3cm}p{1.8cm}p{2cm}p{5cm}p{1.5cm}p{4cm}p{0.5cm}}
  \toprule
   Ref. & Model & Dataset & Description & BP & Limitations & PLA  \\ \hline
   \toprule
   % Action recognition
    \cite{doshi2022federated} & ResNet 8 &AI City Challenge, StateFarm& Optimise FL for resource-limited devices with FedGKT & accuracy = 95\% &  FL results relatively close to the centralized results but are not superior.  &  No\\
   \cite{zhao2020semi} & LSTM & OPP, DG, PAMAP2 & Addressed the problem of the lack of labeled data on the client's level& accuracy = 82\% & Require labeled data on the central server. &  No \\

   \cite{zhang2021driver} & MobileNetV2, ResNet50, VGG16, Xception & StateFarm & Evaluation of FL for driver's action recognition & accuracy = 85\% & Evaluation considered only the accuracy of the model.  &  No\\
   \cite{jain2021federated} & ReNet &HMDB51, UCF101& Custom model initialisation through knowledge distillation with asynchronous aggregation & accuracy = 89.5\% & Limited consideration of non-iid data.  &  No \\ 
   %Crowd counting
   \cite{jiang2020federated} & CNN &MNIST, Next-Character-Prediction & Federated crowd sensing with an incentive mechanism to reward
and motivate participants & accuracy = 80\% & Even though the loss reduced significantly, it is still high.  &  No\\ 
\cite{bao2019flchain} & DDCBF &NA& Suggest an FL framework to distribute trust and incentive among trainers & accuracy = 97\% & Require long time for convergence& No\\
   \cite{shi2018crowd} & DRFL & Wider Face & Proposes a cascade network with two stages trained with FL & mAP = 98.5\% & Can not achieve real-time detection due to computational complexity.  &  Yes\tablefootnote{\url{https://github.com/shizenglin/Deep-NCL}} \\ 
   %Anomaly detection use non CV dataset \cite{singh2021anomaly} 
   \cite{bharti2022edge} & CNN & MVTech & FL-based approach to enable visual inspectors to recognise unseen defects in industrial setups & F1-score >= 90\% & Limited evaluation considering only detection performance.  &  Yes\tablefootnote{\url{https://developer.apple.com/documentation/vision} , \url{https://github.com/hollance/coreml-training}}\\
   \bottomrule
    \end{tabular}
   
\end{table*}

% \cite{} 

% \cite{} 

\subsection{Healthcare and medical AI}
\begin{figure*}[t!]
\begin{center}
\includegraphics[width=1\textwidth]{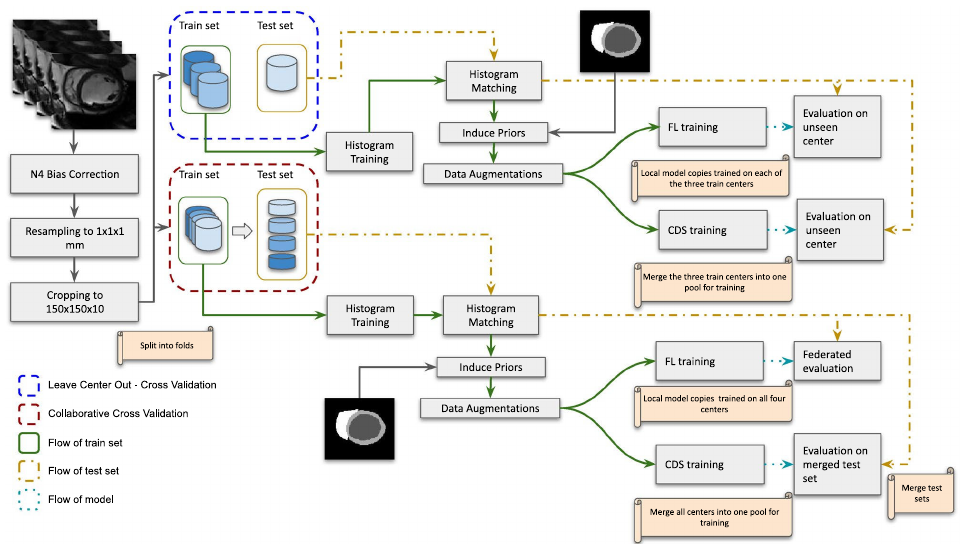}\\
\end{center}
\caption{The methodology applied for conducting the experiments in this study is outlined in \cite{linardos2021federated}. 
Leave center out and collaborative cross-validations,  both employed collaborative data sharing (CDS) and FL for the training.}
\label{flcl}
\end{figure*}

Nowadays, DL methods with large-scale datasets can produce clinically useful models for computer-aided diagnosis \cite{esteva2021deep}. However, privacy and ethical concerns are increasingly critical, which makes it difficult to collect large quantities of data from multiple institutions. FL provides a promising decentralized solution to CL by exchanging client models instead of private data. Sheller et al. \cite{sheller2018multi} performed the first study that investigated the use of FL for multi-institutional collaboration, and enabled the training of DL models without sharing patients' data. In particular, the aggregation process involves calculating a weighted average of institutional updates, where each institution's weight is determined by the proportion of total data instances it holds. This entire process, comprising local training, update aggregation, and distribution of new parameters, is referred to as a federated round. Linardos et al. \cite{linardos2021federated} modeled cardiovascular magnetic resonance using a FL scheme that concentrated on the diagnosis of hypertrophic cardiomyopathy. A 3D-CNN model, pre-trained on action recognition, was deployed. Moreover, shape prior information has been integrated into 3D-CNN using two techniques and four data augmentation strategies (Fig. \ref{flcl}). This approach has then been evaluated on the automatic cardiac diagnosis challenge (ACDC) dataset. The multi-site fMRI classification problem is addressed by \cite{li2020multi} while preserving privacy using a FL model. Accordingly, a decentralized iterative optimization has been deployed before using a randomization mechanism to alter shared local model weights. In the same way, Dayan et al. \cite{dayan2021federated} developed and trained a FL model on data from data from 20 institutes worldwide, namely EMR chest X-ray AI model (EXAM). The latter was built based on the COV-2 clinical decision support (CDS) model. This helps predict the future oxygen requirements of COVID-19 patients using chest X-rays, laboratory data, and inputs of vital signs. In the same way, an FL-based solution that screens COVID-19 from chest X-ray (CXR) images is deployed by \cite{feki2021federated}.  
In addition, a communication-efficient CNN-based FL scheme to multi-chest diseases classification from CXR images is proposed by \cite{cetinkaya2021communication}. Yan et al. \cite{yan2021fedcm} proposed a real-time contribution measurement approach for participants in FL, which is called Fedcm. The latter has been applied to identify COVID-19 based on medical images. Moving forward, Roth et al. \cite{roth2020federated} deployed a FL approach for building medical imaging breast density classification solutions. Data from seven clinical institutions around the world has been used to train the FL algorithm. 

Because of the gaps in fMRI distributions from distinct sites, a mixture of experts domain adaptation (MoE-DA) and adversarial domain alignment (ADA) schemes have been integrated into the FL algorithm. The reference  \cite{yan2020variation} introduced a variation-aware FL approach, in which the variations between clients have been reduced by transforming the images of all clients onto a common image space. A privacy-preserving generative adversarial network, namely PPWGAN-GP is introduced. Moving on, a modified CycleGAN is deployed for every client to transfer its raw images to the target image space defined by the shared synthesized images.  Accordingly, this helps address the cross-client variation problem while preserving privacy. Similarly, for data privacy-preserving, \cite{silva2019federated} used a FL scheme to securely access and meta-analyze biomedical data without sharing individual information. Specifically, brain structural relationships are investigated across clinical cohorts and diseases. Sheller and their colleagues  \cite{sheller2020federated} deployed a FL scheme to facilitate multi-institutional collaborations without the need to share patients' data. In \cite{guo2021multi}, MR data from multiple institutions is shared with privacy preservation. Moreover, cross-site modeling for MR image reconstruction is introduced to reduce domain shift and improve the generalization of the FL model. Moving on, 
\cite{lu2022federated} combined differential privacy and weakly-supervised attention multiple instance learning (WS-AMIL) in order to develop a privacy-preserving FL approach for gigapixel whole slide images in computational pathology. Researchers in 
\cite{li2019privacy} implemented differential-privacy schemes for protecting patients' data in a FL setup designed for brain tumour segmentation on the BraTS dataset. In the same field, Bercea et al. \cite{bercea2021feddis} developed a disentangled FL approach to segment brain pathologies in an unsupervised mode. Cetinkaya et al. \cite{cetinkaya2021improving} attempted to improve the performance of FL-based medical image analysis in non IID settings using image augmentation.

The schemes \cite{han2021application, tekgul2021waffle, liu2021secure} consider whether the necessity of watermarking is required when using FL. For example,  \cite{han2021application} introduced a FL-based zero-watermarking technique for security and privacy preservation in teledermatology healthcare frameworks. FL-based autoencoder is employed for extracting image features from dermatology data using two-dimensional discrete cosine transform (2D-DCT). Conversely, \cite{tekgul2021waffle,liu2021secure} are two strategies proposed for watermarking-based FL. The first is called WAFFLE proposed to prevent global FL model theft by offering a mechanism for model owners to showcase their ownership of the models, and the second is a client-side backdoor-triggered watermarking is adopted to secure FL model verification. Blockchain is also another means used for data privacy, Polap et al. \cite{polap2021agent} developed an intelligent medical system based on agent architecture using blockchain and FL technologies. Since FL algorithms do not inherently contain privacy-preserving mechanisms and can be sensitive to privacy-centered attacks that can divulge patients' data, it is important to augment them with privacy-enhancing technologies, especially in clinical applications that are implemented in a multi-institutional setting. In this context, a differentially private FL solution is suggested by \cite{ziller2021differentially} to segment multi-site medical images in order to further enhance privacy against attacks. Besides, Guo et al. \cite{guo2021towards} propose an FL-based approach for distributed data in medical cyber-physical systems. Their approach helps in training DL models for disease diagnosis following three steps which are repeated in cycle: i) training a global model using offline medical images and transferring the global model to the local diagnosis nodes, ii) re-training the local models using local data and iii) sending them back to the central server for federated averaging. The authors in \cite{jiang2021harmofl} propose a technique to harmonize local and global drifts in FL models on heterogeneous medical images.
In doing so, the local update drift is first mitigated by normalizing amplitudes of images transformed into the frequency domain and then a client weight perturbation guiding each local model to reach a flat optimum is designed based on harmonized features.

%EXAM achieved an average area under the curve (AUC) >0.92 for predicting outcomes at 24 and 72 h from the time of initial presentation to the emergency room, and it provided 16\% improvement in average AUC measured across all participating sites and an average increase in generalizability of 38\% when compared with models trained at a single site using that site’s data. For prediction of mechanical ventilation treatment or death at 24 h at the largest independent test site, EXAM achieved a sensitivity of 0.950 and specificity of 0.882. In this study, FL facilitated rapid data science collaboration without data exchange and generated a model that generalized across heterogeneous, unharmonized datasets for prediction of clinical outcomes in patients with COVID-19, setting the stage for the broader use of FL in healthcare. 

The problem setting of federated domain generalization (FedDG) is solved in \cite{liu2021feddg}. This helps in learning a FL architecture from various distributed SDs, which enables its generalization to unseen TDs. This was made possible by introducing a the episodic learning in continuous frequency space (ELCFS) technique. Moving on, \cite{sun2021partialfed} propose a partial initialization-based cross-domain personalized FL, namely PartialFed. Their method is based on loading a subset of the global model's parameters instead of loading the entire model, as it is done in most of the FL approaches. Thus, it is closer to the split Learning paradigm. More over, Wang et al. \cite{wang2021progfed} design an effective communication, and computation efficient FL scheme using progressive training. This helps to reduce computation and two-way communication costs, while preserving almost the same performance of the final models. While FL allows collaboratively training, using a joint model that is trained in multiple medical centers that maintain their data decentralized to preserve privacy, the federated optimizations face the heterogeneity and non-uniformity of data distribution across medical centers. To overcome this issue, an FL scheme with shared label distribution, namely FedSLD, is proposed by \cite{luo2021fedsld}. This approach can reduce the discrepancy brought by data heterogeneity and adjust the contribution of every sample to the local objective during optimization via the knowledge of clients' label distributions. 
Tbale \ref{tab:FL2} presents a Summary of FL frameworks proposed for healthcare and AI applications.

\begin{table*}[t!]
\caption{Summary of FL frameworks proposed for healthcare and medical AI applications. Best performance (BP), project link availability (PLA).} 
\label{tab:FL2}

\renewcommand{\arraystretch}{1.5} % Reset to original spacing

\begin{tabular}{p{0.5cm}p{1.5cm}p{2.5cm}p{4.5cm}p{1cm}p{4.6cm}p{0.5cm}}
  \toprule
   Ref. & Model & Dataset & Description & BP (\%) & Limitations & PLA   \\ \hline
   \toprule
\cite{linardos2021federated} & 3D-CNN & M\&M and ACDC datasets & Modeling cardiovascular magnetic resonance with focus on diagnosing hypertrophic cardiomyopathy & AUC = 78 & Presence of bias against ACDC on the shape and intensity set-up where FL exhibits an AUC performance of about 0.85 to 0.89. &  Yes\tablefootnote{\url{https://github.com/Linardos/federated-HCM-diagnosis}} \\

\cite{chen2021personalized} & PFA & Real-World Dermoscopic FL Dataset \tablefootnote{\url{https://github.com/CityU-AIM-Group/PRR-FL}} & A personalized retrogress-resilient FL with modification in the clients and server.  & AUC= 88.92, F1= 70.75 & Generalization with  other datasets is not confirmed. &  Yes\tablefootnote{\url{https://github.com/CityU-AIM-Group/PRR-FL}}\\

\cite{li2020multi} & MoE-DA, ADA &  Autism Brain Imaging Data Exchange dataset (ABIDE I) & Privacy-preserving FL and domain adaptation for multi-site fMRI analysis  & ACC= 78.9 & The model updating strategy is not optima. Additionally,the sensitivity of the mapping function was difficult to estimate. &  Yes\tablefootnote{\url{https://github.com/xxlya/Fed_ABIDE}}\\

\cite{yan2020variation} & PPWGAN-GP, modified CycleGAN & LocalPCa, PROSTATEx challenge \cite{litjens2014computer} & Address cross-client variation problem among medical image data using VAFL. & AUC= 96.79, ACC=  98.3\%& Do no address the inter-observer problem and  incomplete image-to-image translation. &  No  \\

\cite{silva2019federated} & fPCA & ADNI, PPMI, MIRIAD and UK Biobank & Meta-analysis of large-scale subcortical brain data using FL & N/A & No comparisons with the SOTA have been reported. &  No\\

\cite{dayan2021federated} & fPCA & Synthetic and real world private data & Predict the future oxygen requirements of symptomatic patients with COVID-19 using inputs of vital signs & AUC> 92 & Normalization techniques to enable the training of AI models in FL were not investigated. &  Yes\tablefootnote{\url{https://ngc.nvidia.com/catalog/models?orderBy=scoreDESC&pageNumber=0&query=covid&quickFilter=models&filters}} \\

\cite{polap2021agent} & Agent-based mod & Private data & Blockchain technology and threaded FL for private sharing of medical images. & ACC= 80  & Poor internet connection hinders accessing the data. The initial adaptation of the
classifier for practical use needs more investigation. &  No\\

\cite{luo2021fedsld} & FedSLD  & MNIST, CIFAR10,  OrganMNIST(axial), PathMNIST& FL with shared label distribution to classify medical images & ACC= 95.85 & Reduce the impact of non-IID data by leveraging the clients' label distribution. &  No \\

\cite{sun2021partialfed} & PartialFed, PartialFed-Adaptive& KITTI, WFace, VOC, LISA, DOTA, COCO, WC, CP, CM, Kit, DL & Partial initialization for cross-domain personalized FL & ACC= 95.92 & Reduce performance degradation caused by extreme distribution heterogeneity. &  No \\
\bottomrule

\end{tabular}
   
\end{table*}

\subsection{Autonomous driving}
Autonomous driving has recently received increasing interest due to the advance of CV, which is in the core of this technology. Vehicles leverage object detectors to analyze images collected by multiple sensors and cameras, analyze their surroundings in real-time and then recognize different objects, including other vehicles, road signs, barriers, pedestrians, etc., which help them in safely navigating the roads.
While a plethora of studies have focused on improving the accuracy by training DL algorithms on centralized large-scale datasets, few of them have addressed the users' privacy. To that end, using FL in autonomous driving has recently attracted increasing attention. However, numerous challenges have been raised, including data discrepancy across clients and the server, expensive communication, systems heterogeneity and privacy concerns. Typically, privacy issues include internal and external data, such as the faces of pedestrians, vehicles' positions, etc. 
%The capabilities, sensory equipment, and the driving environment of the agents introduce variety in the distributions of the collected data and do not allow for the assumption of being independent and identically distributed. Lastly, there is still an open problem of ensuring fairness in incorporating the model updates collected through on-device learning.

\begin{figure*}[t!]
\centering
\includegraphics[width=0.9\textwidth]{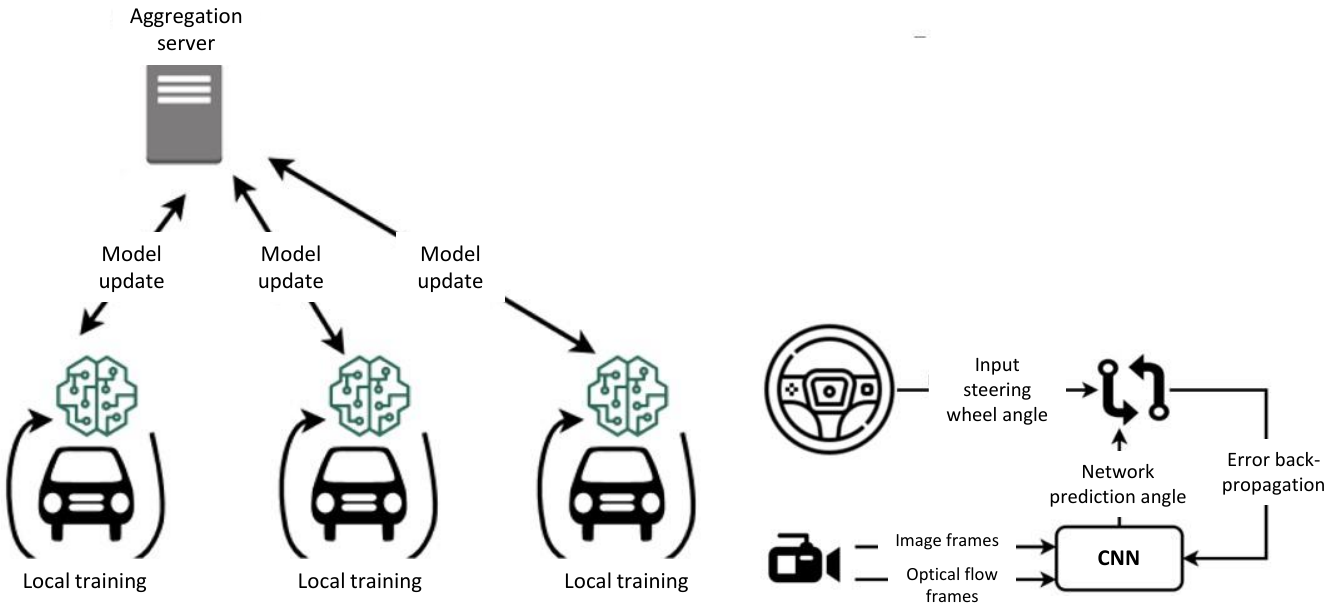}
\caption{FL-based framework for wheel steering angle prediction in autonomous vehicles \cite{zhang2021end}.}
\label{fi:FL-autonomous}
\end{figure*} 

 The approach to on-device ML using FL and validates it through a case study on wheel steering angle prediction for autonomous driving vehicles is presented in \cite{zhang2021end}. The results show that FL can significantly improve the quality of local edge models and reach the same accuracy level as centralized ML without negative effects. FL can also accelerate model training speed and reduce communication overhead, making it useful for deploying ML/DL components to various embedded systems. Fig. \ref{fi:FL-autonomous} presents a FL-based framework for wheel steering angle prediction in autonomous vehicles. Nguyen et al.
\cite{nguyen2022deep} propose a communication-efficient FL to detect fatigue driving behaviors, namely FedSup. It helps in progressively optimizing the sharing model with tailored client–edge–cloud architecture and reduces communication overhead by a Bayesian convolutional neural network (BCNN) data selection strategy. In \cite{nguyen2022deep}, a federated autonomous driving network (FADNet) solution is introduced for enhancing models' stability, ensuring convergence, and handling imbalanced data distribution problems where FL models are trained. 

\begin{figure*}[t!]
\centering
\includegraphics[width=0.9\textwidth]{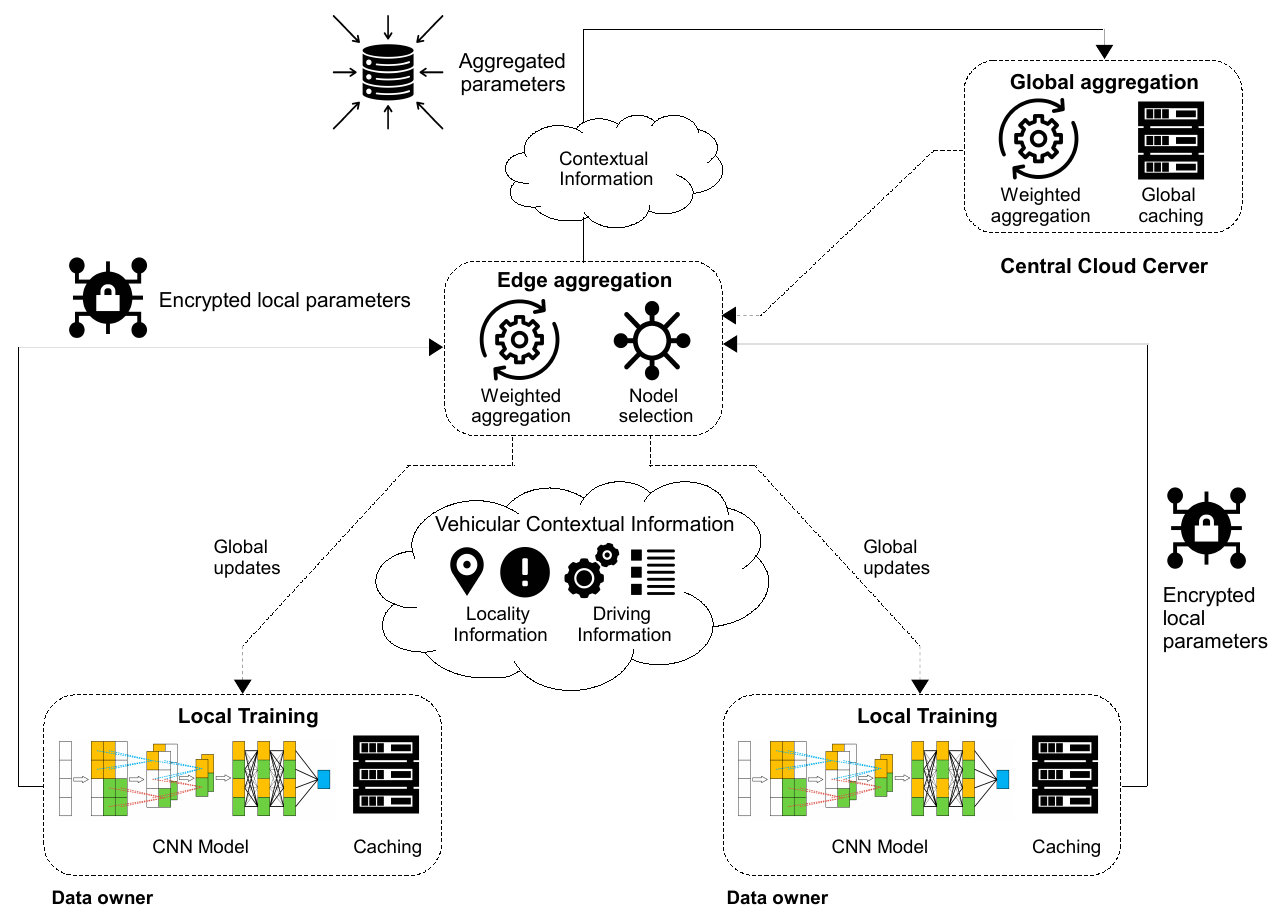}
\caption{Overview of the the TFL-CNN framework introduced in \cite{zhou2021two}.}
\label{fig:CFL}
\end{figure*}

Zhou and their colleagues \cite{zhou2021two} discuss the need for distributed ML techniques to take advantage of the massive interconnected networks and heterogeneous data generated at the network edge in the upcoming 6G environment. A two-layer FL model is proposed that utilizes the distributed end-edge-cloud architecture to achieve more efficient and accurate learning while ensuring data privacy protection and reducing communication overhead. A novel multi-layer heterogeneous model selection and aggregation scheme is designed to better utilize the local and global contexts of individual vehicles and roadside units (RSUs) in 6 G-supported vehicular networks. This context-aware distributed learning mechanism is then applied to address intelligent object detection in modern intelligent transportation systems with autonomous vehicles. Fig. \ref{fig:CFL} presents an overview of the two-layer FL model based on convolutional neural network (TFL-CNN) framework.
Table \ref{tab:FL-vehicles} presents a summary of FL frameworks proposed for autonomous vehicles.

\begin{table*}[t!]
\caption{Summary of FL frameworks proposed for autonomous vehicles, including the ML model used, dataset, description of the main contribution,best performance (BP), limitations and project link availability (PLA).} \label{tab:FL-vehicles}

\renewcommand{\arraystretch}{1.6} % Reset to original spacing

\begin{tabular}{p{0.5cm}p{1.5cm}p{2.3cm}p{4.5cm}p{1.8cm}p{4.16cm}p{0.6cm}}
  \toprule
   Work & Model & Dataset & Description & BP & Limitations & PLA \\ \hline
   \toprule
   \cite{nguyen2022deep} & FADNet & Udacity, Gazebo, Carla & P2P federated framework for training autonomous driving models & RMSE= 0.07 & Limited deployment experiment &  Yes\tablefootnote{https://github.com/aioz-ai/FADNet} \\
   \cite{khan2022dispersed} & BSUM-based solution & MNIST & Dispersed FL framework to improve the robustness, privacy-awareness and communication constraints & accuracy = 99\% & NP-hard and non-convex formulation of the problem & No \\
   \cite{li2021privacy} & CNN & Private dataset &Addresses the case where vehicles and servers are considered honest but curious through blockchain & accuracy= 92.5\% & Require long training iterations for small images & No\\
   \cite{donevski2021addressing} & CNN & MNIST, FMNIST & A reactive method for the allocation of wireless resources, dynamically occurring at each round & accuracy = 88 \% & The impact of resource allocation methods is diminished when
strengthening the role of the proximal term & No\\
   \cite{zhang2021end} & CNN & SullyChen &End-to-end ML model for handling real-time generated data & RMSE= 9.2 & Synchroneous aggregation &  No\\
   \cite{elbir2022federated} & U-NET, CNN & Lyft Level 5 AV &Feasibility study of using FL for vehicular applications &  accuracy = 95\% & Evaluated only the accuracy & No \\
   \cite{zhou2021two} & TFL-CNN & BelgiumTSC & An improved model selection and aggregation algorithm & F1-score = 94\% & Limited evaluation of the framework & No \\
   \cite{jallepalli2021federated} & YOLO3 & KITTI & Evaluation of real-time object detection in  real traffic environments & MAP = 68.5\% & Sensitive to changes in labels across clients & No \\
   \cite{xie2022efficient} & SNNs & BelgiumTSC & Leverages spike NN to optimize resources required for the training & accuracy = 95\% & Suffers from security issue & No \\
   \bottomrule
    \end{tabular}
\end{table*}

\section{Open issues}
\label{issues}
The advances in CV algorithms have increased the number and variety of tasks that range from the detection and tracking of specific objects in static images or controlled image streams to public surveillance and the detection of normal or abnormal behaviors and conditions in the wild, with applications from agricultural crop monitoring to law enforcement and medical diagnostics. The information captured in surveillance cameras' footage may contain sensitive and private data that must be protected (e.g., the presence of individuals at a place), the CV models trained to detect abnormal behaviors in public may suffer from race, ethnicity or gender bias, and scene perception software in self-driving cars may suffer from the ability to adapt to new environments or conditions (e.g. fog, mist or haze).

FL is a new method for protecting privacy when developing a DNN model, that uses data from various clients and trains a common model for all clients. This is achieved by fusing distributed ML, encryption and security, and introducing incentive mechanisms based on game theory and economic theory. FL could therefore serve as the cornerstone of next-generation ML that meets societal and technological requirements for ethical AI development and implementation. 

Despite their many advantages, FL solutions also have several challenges to address. First of all, since FL is a distributed learning technique it is important to guarantee the efficient communication between the federated network nodes so that they can communicate the learned model parameters. In addition to this, the models may be trained on nodes of different hardware architecture and capabilities (e.g. IoT devices, smartphones, etc.) which constitute a heterogeneous environment for FL. Consequently, it is important to establish the mechanisms that manage heterogeneous nodes in the same network, considering their restrictions and capabilities. The use of central (cloud-based) nodes that carry part of the training process, either using additional training data or by training parts of the learning model (e.g. some layers of the DNN as in the split learning paradigm) can be beneficial for the overall performance. A third issue that must be considered is the statistical heterogeneity of data that arrives in each node. Learning using non-IID data is still an open challenge for FL algorithms, and various alternatives are still under evaluation. Finally, privacy and robustness concerns are still present and methods that preserve privacy and at the same time guarantee the resulting model robustness to any kind of breaches or attacks must be properly designed and developed.

\subsection{Communication overhead} 

A crucial component of a FL environment is communication between the client nodes and the central server. The ML model is downloaded, uploaded, and trained over the course of several communication rounds. Although transferring models instead of training datasets is significantly more efficient, these communication rounds may be delayed when the device has limited bandwidth, energy, or power. The communication overhead also increases when multiple client devices participate in each communication round thus leading to a bottleneck. Further delays may occur from the need to synchronise the receipt of models from all clients, including those with low bandwidth or unstable connections. An additional burden to the client synchronization overhead can be caused by the non-identical distribution of training data to the nodes, which may delay the training of some models, and consequently the model update process \cite{sattler2019robust}. 
This delay in cybersecurity can be considered an intrusion if it surpasses a predetermined threshold \cite{kheddar2023deep}.

In order to mitigate the communication overhead and establish communication-efficient federation strategies, \cite{haddadpour2021federated} have proposed the compression of transferred data. In a different approach, \cite{luping2019cmfl} have focused on identifying irrelevant models and precluding them from the aggregation, thus significantly reducing the communication cost. This is achieved by sharing the global tendency of model updating between the central server and all the client nodes and asking each client to align its update with this tendency before communicating the model.

In the case of FLchain \cite{nguyen2021federated}, the model communication cost between the clients and a central server, which is common in FL, is replaced by the cost of sharing the models with the blockchain ledger. For this reason, it is important to consider the time needed for this process, including local training, model transmission, consensus, and block mining.

\subsection{Heterogeneity of client nodes} 
One of the major challenges in FL is the heterogeneity of data on different clients, which can hinder effective training. To address this issue, client selection strategies are often used in an attempt to improve the convergence rate of the FL process. While active client selection strategies have been proposed in recent studies, they do not take into account the loss correlations between clients and only offer limited improvement compared to a uniform selection strategy \cite{varlamis2022using}. To overcome this limitation, FedCor, an FL framework that utilizes a correlation-based client selection strategy to improve the convergence rate of FL is proposed in \cite{tang2022fedcor}. The loss correlations between clients is modeled using a Gaussian process (GP) and use this model to select clients in a way that significantly reduces the expected global loss in each round. Moreover, an efficient GP training method is developed with low communication overhead for use in the FL scenario by leveraging the covariance stationarity. The experimental results show that FedCor can improve convergence rates by 34\% to 99\% on FMNIST and 26\% to 51\% on CIFAR-10 compared to the SOTA method.
Besides, FL still suffers from significant challenges such as the lack of convergence and the possibility of catastrophic forgetting.
However, it is demonstrated in \cite{qu2022rethinking} that the self-attention-based architectures (e.g., Transformers) have shown to be more resistant to changes in data distribution and therefore can improve the effectiveness of FL on heterogeneous devices.
The authors of \cite{mendieta2022local} propose an alternative approach, called FedAlign, which aims to address data heterogeneity by focusing on local learning rather than proximal restriction. They conducted a study using second-order indicators to evaluate the effectiveness of different algorithms in FL and found that standard regularization methods performed well in mitigating the effects of data heterogeneity. FedAlign was found to be a simple and effective method for overcoming data heterogeneity, with competitive accuracy compared to SOTA FL methods and minimal computational and memory overhead.

On the other hand, when FL is limited only to client nodes that share the same model architecture then the FedAvg algorithm and its alternatives (e.g. FedSDG, FedProx, etc.) can be applied to merge the locally trained models in each iteration \cite{bousbiat2023crossing}. However, such approaches assume that the underlying nodes share similar hardware architecture, specifications and processing capabilities in general, which is not always the case in FL settings. For example, when smartphones, monitoring cameras and field programmable gate arrays (FPGA) attached cameras are orchestrated in an FL setting, the models have heterogeneous architectures and model parameter sharing is infeasible \cite{dominguez2019stereo,azzouzi2023novel}. The barriers of conventional FL are removed when the models share their models instead of their parameters or updates. Algorithms such as the federated model distillation (FedMD) propose a model agnostic federation solution, which sends the predictions of the local models to the central node instead of the models themselves \cite{li2019fedmd}.

Another impact of node heterogeneity applies to the overall performance of the FL process since the federation of heterogeneous nodes with varying training data structure and size and varying processing capabilities usually requires in each round all the local nodes to train their models before proceeding to the next iteration. As a result, slow nodes, with low usability data may degrade the time performance of the federation. The work \cite{pang2020realizing} proposes a reinforcement learning-based central server, which gradually weights the clients based on the quality of their models and their overall response in an attempt to establish a group of clients that are used to achieve almost optimal performance.

\subsection{Non-IID data} 
Although FL offers a promising method to privacy protection, there are significant difficulties when FL is used in the real world as opposed to centralized learning.  Numerous studies have shown that the accuracy of FL on non-IID or heterogeneous data would inevitably deteriorate, mainly because of the divergence in the weights of local models that result from non-IID data \cite{alsalemi2021smart}. Either the FL approach is horizontal (i.e. aggregating the local models' weights on a central server) or vertical (i.e. aggregating the model outputs in the guest client to calculate the loss function) the non-IID data can take various forms (e.g. attribute, label or temporal skew) that can affect the overall performance, and mitigation measures must be taken to avoid it \cite{zhu2021federated}. These include data sharing and augmentation, and the fine-tuning of local models using a combination of local and global information. Wang wt al. \cite{wang2020optimizing} propose FAVOR, a reinforcement-based method (deep Q-learning) for choosing the clients that contribute models to the aggregation phase in each round. The proposed method reduces the non-IID data bias and the communication overhead. Since the number of dimensions of the state space equals the number of model weights times the number of client nodes, a dimensionality reduction technique, such as PCA, is applied in order to compress the state space. Another group of approaches tries to balance the bias of non-IID data by clustering the local updates using a hierarchical algorithm  \cite{briggs2020federated}. Such approaches result in multiple models that are trained independently and in parallel, leading to a faster and better convergence of each group.

\subsection{Device compatibility and network issues} 
FL relies on the participation of multiple devices, which may have different hardware and software configurations. This can make it difficult to ensure that the model can be trained on all devices, and may require additional efforts to optimize the model for different device types \cite{bousbiat2023neural}.
FL requires a stable and reliable network connection in order to train the model and exchange model updates between participants and the central server. If the network connection is unreliable or has low bandwidth, this can negatively impact the training process and the performance of the model.
Explicitly, if the parties involved in FL are located in different geographical regions, the time it takes for model updates to be transmitted over the network can vary significantly, leading to delays in the training process.

\subsection{Human bias} 
FL relies on the participation of multiple parties, which may have different biases or perspectives that can influence the model's training and performance. This can be particularly challenging in the context of CV, as the model may be trained on data with biased or inaccurate annotations. It is important to carefully consider and address these issues in order to ensure that the model is fair and unbiased \cite{pouriyeh2022secure}.

\subsection{Privacy and robustness to attacks} 
The resilience and data privacy of existing FL protocol designs have been shown to be compromised by adversaries both inside and outside the system. The sharing of gradients during training can reveal private information and cause data leakage to the central server or a third party. Similarly, malicious users can try to affect the global model by introducing poisoned local data or gradients (model poisoning) in an attempt to destroy its convergence (i.e. Byzantine attack) or to implant a trigger in the model that constantly predicts adversarial classes, without losing performance on the main task. These two times of attacks either aim to breach user privacy (e.g. infer class representatives, infer class membership or infer user properties) or poison the model in order to control its behavior \cite{lyu2020privacy}.
Homomorphic encryption, secure joint computation from multiple parties, and differential privacy are some of the means for mitigating privacy breaches. Respectively, the defenses against poisoning attacks focus on the detection of malicious users based on the anomalies detected in their models (e.g. different distributions of features than the rest of the users \cite{shen2016auror}) focusing on increasing the robustness against Byzantine attacks. They also examine the resulting models to detect whether they are compromised or not, using a combination of backdoor-enabled and clean inputs and examining the behavior of the model \cite{andreina2021baffle}.

Blockchain can be beneficial for preserving privacy in FL settings and can help avoid various attack types, from single point of failure and membership inference to Byzantine, label flipping, and data poisoning ones \cite{qi2021privacy}. The FL blockchain, or FLchain for short, paradigm can transform mobile-edge computing to a secure and privacy-preserving system, once the proper solutions are found for allocating resources, providing incentives to the client nodes, and protecting the security and privacy of data at an optimal communication cost.
In their proposed architecture, the mobile-edge computing (MEC) servers can act either as learning clients or as miners to establish blockchain consensus and the mobile devices associated with each server focus only on the learning tasks. The mobile devices transmit their models to the MEC server as transaction. Each MEC server stores the local models it collects to the blockchain after verifying them with other MEC servers. The aggregation node or any other local node is then able to retrieve the stored and verified models from the blockchain, aggregate them and use the resulting model in the next iteration \cite{nguyen2021federated}. Watermarking is another means of model security preserving \cite{han2021application, tekgul2021waffle, liu2021secure}.

%\subsection{Performance}
%\textcolor{black}{@hamza: the information of this subsection  is repeated elsewhere in the open issue introduction.}
%FL may not always achieve the same level of performance as traditional centralized ML, due to the challenges of training on decentralized and heterogeneous data. In addition, the communication overhead and other technical challenges of FL may limit the speed and efficiency of training.

\subsection{Replication of research results}
\begin{figure*}[ht!]
\begin{center}
\includegraphics[width=1\textwidth]{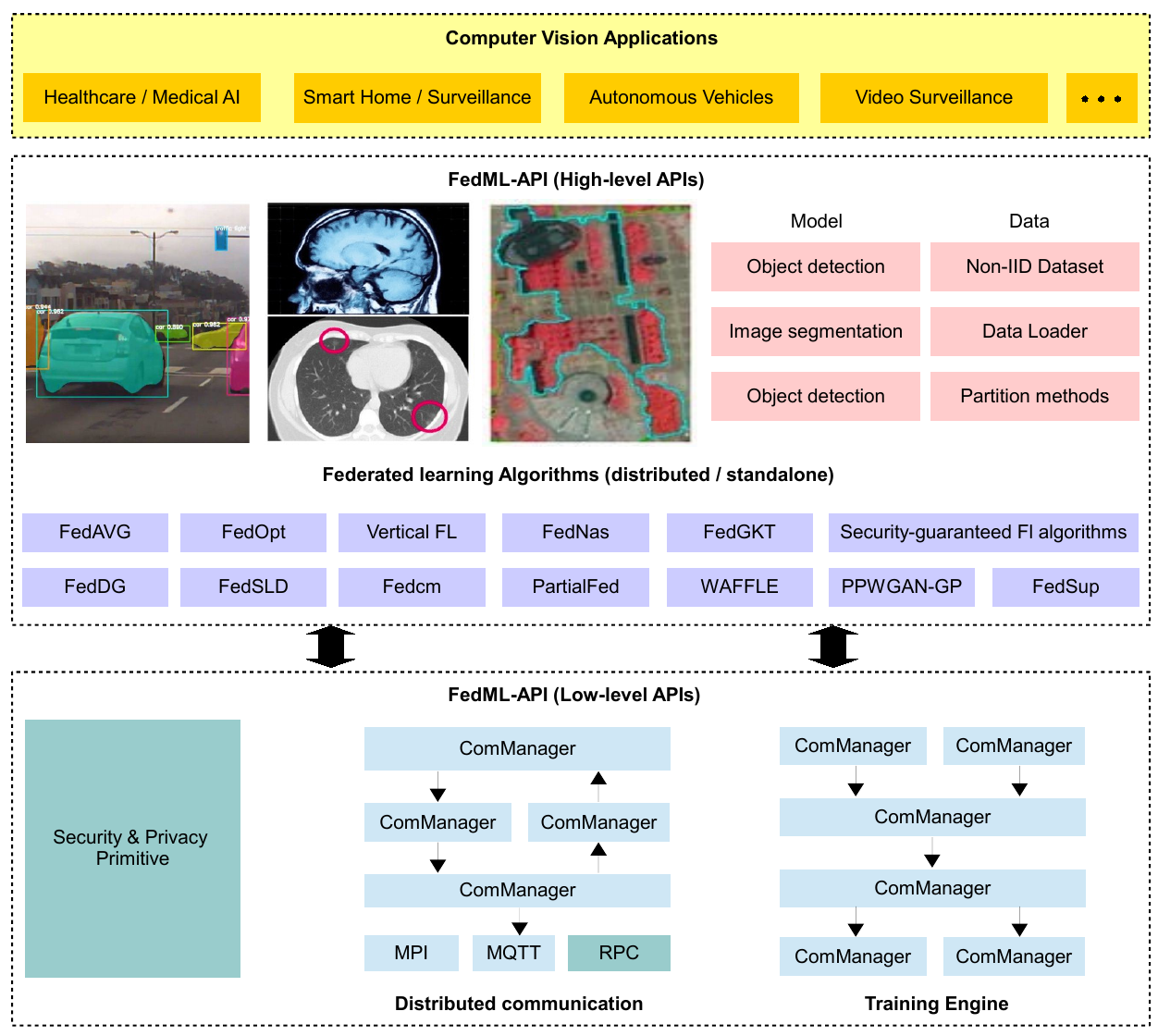}\\
\end{center}
\caption{Overview of FedCV platform architecture design proposed in \cite{he2021fedcv}.}
\label{fig:FedCV}
\end{figure*}

Despite the fact that the performance of FL models requires further improvement compared to that obtained with centralized training, replicating FL research results and conducting fair comparisons are still challenging. This is mainly due to the lack of exploration in different tasks with a unified FL framework. To that end,  \cite{he2021fedcv} develop an FL benchmarking and library platform for CV applications, called FedCV. This helps in bridging gaps between SOTA algorithms and facilitating the development of FL solutions. 
Additionally, three main taks of CV, including image object detection, image segmentation and image classification can be evaluated on this toolkit, as shown in Fig. \ref{fig:FedCV}. Moreover, numerous FL algorithms, models, and non-IID benchmarking datasets have been uploaded and still the toolkit can be updated.

Besides, enhancing FL-based systems' efficiency is a delicate task due to the per-client memory cost and large number of parameters. Thus, using the FedCV framework, which is an efficient and flexible distributed training toolkit that has easy-to-use APIs, along with benchmarks, and different evaluation settings can be useful for the FL research community to conduct advanced research CV studies. 

\section{Challenges of using FL in CV}
FL is a promising approach for solving privacy and data distribution challenges in CV tasks such as object detection, autonomous vehicles, etc. However, there are several challenges associated with the use of FL in these applications. For instance, FL assumes that the data distribution across clients is similar. However, in CV tasks, the data may be heterogeneous, and the models need to be robust to handle variations in lighting, viewpoint, and occlusions. FL relies on communication between clients and the server. For CV tasks, such as object detection, the models can be large, and the communication costs can be high. This can result in slower training times and higher energy consumption.
Moreover, CV tasks require labeled data for training the models while FL requires each client to have labeled data, which can be challenging in scenarios where labeling data is expensive or time-consuming.
Additionally, FL requires clients to share their data with the server, which can raise privacy and security concerns. Malicious clients can manipulate the data or models, compromising the integrity of the system.
Lastly, FL requires aggregating the models from multiple clients to produce a global model. In CV tasks, this can be challenging due to differences in the quality of the models and the data distribution across clients.

%On the other hand, when an object detection model is trained by a single party (server) and distributed to other users (clients), it may suffer from significant performance degradation. This can be particularly evident in autonomous driving situations, where various driving environments can result in noticeable domain shifts, causing the model to produce biased predictions. However, FL allows for collaborative training across multiple parties without compromising the privacy of client data.

On the other hand, FedCV \cite{he2021fedcv} is a benchmarking framework for the evaluation of FL in popular CV tasks such as image classification and segmentation and object detection. It comprises non-IID datasets  and various models and algorithms for experimentation. The experiments validate the aforementioned open challenges for FL in CV tasks: i) the degradation of model accuracy when non-IID data are used, ii) the complexity of optimising the training in the FL setting, iii) the huge number of parameters of NN models used in CV that affects the FL performance. The same benchmark can be employed to validate all the remaining issues discussed in this section, such as the node and data heterogeneity, the need for robustness and privacy etc.

FedVision \cite{liu2020fedvision} is an online platform for developing object detection solutions using FL and a three step workflow that comprises: i) image annotation, ii) horizontal FL model training and iii) model update. Since the platform is generic, it allows users to configure the learning parameters, schedule the communication between the server and clients, allocate tasks and monitor the utilization of resources. The main challenges mentioned in this section for FL also apply in the case of FedVision, which however chooses specific strategies to tackle them. For example, it uses model compression to reduce communication overhead, cloud object storage to store huge amounts of data (model parameters) in the server, and an one-stage approach, based on YOLOv3, to perform end-to-end training of the model that identifies the bounding box and the class of the object. However, the real challenges for FL approaches in CV come with the application in real-world images from surveillance cameras, etc. \cite{luo2019real}. The large-scale of data collected and the requirement for almost real-time inference raises more design challenges for FL experts.

\section{Future directions}
\label{future}
The future of FL is very promising, but also challenging for researchers, with the main directions being as follows:
\begin{itemize}
    \item \textbf{Deployment over heterogeneous environments:} Smartphones are becoming the most popular edge devices for ML applications since they allow users to perform a large variety of tasks from face detection to voice recognition. In this direction, FL can be used to support such tasks without exposing private information. On the other side, IoT networks combine wearables, mobile and stable sensors, and smartphones in order to establish smart environments for the end users. All these constitute a diverse and heterogeneous environment in which models of varying complexity have to be communicated in the place of data in order to implement federation while protecting data privacy.
    \item \textbf{Efficient communication:} When creating techniques for federated networks, communication is a crucial bottleneck to take into account. This is due to the fact that federated networks may contain a sizable number of devices (such as millions of smartphones), and communication throughout the network may be much slower than local computing. Reducing the overall number of communication rounds or the quantity of communicated messages at each round are the two main ways to further cut communication. The communication can be more efficient using: i) Local updating techniques that allow to cut down the overall communication rounds, ii) model compression techniques including quantization, subsampling, and sparsification can be used to reduce the size of messages conveyed during each update round, iii) when the connection with the server becomes a bottleneck, decentralized topologies provide an alternative, particularly when working in networks with low bandwidth or excessive latency. Iterative optimization algorithms are parallelized using asynchronous communication. An appealing strategy for reducing stragglers in heterogeneous contexts is the use of asynchronous systems. Moreover,  using 5G/6G networks will offer significantly higher speeds and lower latency compared to previous generations. This can enable faster and more efficient FL in CV applications.
    
    \item \textbf{Dispersed FL:} Concerns about FL's robustness exist since it could cease to function if the aggregate server fails (e.g., due to a malicious attack or physical defect). Dispersed FL can be used as a more robust alternative to FL, with groups of nodes with a lot of computing power to collaborate in more sub-global iterations to increase their performance and consequently support the overall performance of the federation.
    \item \textbf{New organizational models:} The term "devices" can refer to entire companies or institutions in the context of FL. For applications in predictive healthcare, hospitals are an example of a company with a lot of patient data. Hospitals must adhere to strong privacy laws and may have to maintain local data due to ethical, administrative, or regulatory requirements. Such applications can benefit from FL since they can ease network load and enable private learning amongst many devices and organizations.

     \item \textbf{Large language models and generative chatbots:} Advanced language models, such as ChatGPT, can assist in various ways to improve the use of FL in CV by (i) offering explanations, answering questions, and providing tutorials related to FL and CV \cite{sohail2023decoding}. By helping researchers and practitioners understand the concepts better, it can drive wider adoption and more informed application of FL in CV \cite{farhat2023analyzing}; (ii) simplifying complex algorithms, provide pseudocode, and suggest optimization strategies; assisting with code debugging or suggesting modifications to FL algorithms, thereby improving their efficiency or performance \cite{sohail2023future}; (iii) providing insights into the ethical implications and considerations when applying FL in CV, especially in terms of data privacy and usage; (iv) simulating client-server conversations, allowing developers to anticipate challenges and refine their systems; and (v) offering advice on best practices for integration, be it with databases, cloud services, or edge devices \cite{sohail2023using}.
    
\end{itemize}

\section{Conclusion}
\label{conclusion}
Federated Learning (FL) has emerged as a revolutionary paradigm in the realm of Computer Vision (CV), fostering collaborative machine learning without compromising data privacy. This review navigated through the intricate alleys of FL, from its foundational concepts to the myriad applications in CV. The aggregation approaches such as averaging, Progressive Fourier, and FedGKT accentuate FL's versatility. Moreover, the inclusion of privacy technologies like the Secure MPC model, differential privacy, and homomorphic encryption underscores its commitment to safeguarding data.

It is remarkable to note the vast landscape of CV applications benefitting from FL, ranging from object and face detection to innovative domains like healthcare, autonomous driving, and smart environment surveillance. Yet, like any evolving technology, FL in CV is not devoid of challenges. Issues like communication overhead, device heterogeneity, and the conundrums posed by non-IID data offer fertile grounds for future research.

While current advancements set a promising trajectory, the open issues highlight areas ripe for exploration and innovation. The challenges also underscore the importance of collaboration between researchers, practitioners, and industries to make FL more efficient, inclusive, and robust for CV.

As we stand on the cusp of a technological evolution, FL offers a beacon of hope, combining the best of collaborative learning and data privacy. The journey ahead is replete with opportunities and challenges, making it an exhilarating era for researchers and enthusiasts alike.

\end{document}